\documentclass[12pt]{article} 
\usepackage{bm}
\usepackage{graphicx}
\usepackage{fullpage}
\usepackage[auth-lg]{authblk}
\usepackage{hyperref}

\title{Central pattern generators evolved for real-time adaptation to rhythmic stimuli}

\author{Alex Szorkovszky\,$^{1,2,*}$, Frank Veenstra\,$^{2}$ and Kyrre Glette\,$^{1,2}$}
\affil{$^{1}$RITMO Centre for Interdisciplinary Studies in Rhythm, Time and Motion, University of Oslo, Oslo, Norway\\
$^{2}$Department of Informatics, University of Oslo, Oslo, Norway\\
$^*$ \small{Email: alexansz@ifi.uio.no}}

\date{}

\begin{document}

\maketitle

\begin{abstract}
For a robot to be both autonomous and collaborative requires the ability to adapt its movement to a variety of external stimuli, whether these come from humans or other robots. Typically, legged robots have oscillation periods explicitly defined as a control parameter, limiting the adaptability of walking gaits. Here we demonstrate a virtual quadruped robot employing a bio-inspired central pattern generator (CPG) that can spontaneously synchronize its movement to a range of rhythmic stimuli. Multi-objective evolutionary algorithms were used to optimize the variation of movement speed and direction as a function of the brain stem drive and the center of mass control respectively. This was followed by optimization of an additional layer of neurons that filters fluctuating inputs. As a result, a range of CPGs were able to adjust their gait pattern and/or frequency to match the input period. We show how this can be used to facilitate coordinated movement despite differences in morphology, as well as to learn new movement patterns.

\end{abstract}

\section{Introduction}

%Current uses of biological CPGs
Biologically inspired central pattern generators (CPGs) are useful for their properties, typical of self-organized systems, such as distributed control and robustness to perturbations \cite{ijspeert2008central,aoi2017adaptive}. This allows adaptive behaviours such as compensation for physical damage \cite{dasgupta2015distributed} or walking in novel environments \cite{steingrube2010self}.
Spontaneous entrainment of motion patterns to sensory input is also expected from such systems, and adaptation of bio-inspired CPGs to body and environmental mechanics has indeed been widely demonstrated \cite{buchli2006finding,iwasaki2006sensory,seo2010cpg,thandiackal2021emergence}. Likewise, the movement pattern itself can be determined by interactions with the environment \cite{maufroy2010integration,fukuoka2015simple}.

%Need for gait flexibility
While there typically exists a `natural' gait frequency for a particular gait with a particular body (i.e.\ one that maximizes energy efficiency), it is often necessary to modulate one's gait frequency. A cat, for example, may need to walk at a slow pace in order to ambush its prey. For robots, being able to adapt to humans in their vicinity is an important goal, particularly for caring and collaborative applications \cite{nocentini2019survey}. For early humans, it is widely thought that the ability to synchronize movements with others was a crucial step in the evolution of social cognition \cite{knoblich2008evolving,tomlinson2015million,kotz2018evolution}. This is therefore a relevant capability for socially responsive and intelligent robotics.

%Gait flexibility in CPGs
In vertebrates, gait pattern and frequency modulation are indirectly controlled by the intensity of the current from the brain stem, to which the spiking rates of locomotor neuron populations are sensitive \cite{danner2017computational}. Early bio-inspired robotics work attempted to replicate this emergent pattern generation, typically with continuous-time recurrent neural networks \cite{beer1997biologically,ijspeert2001connectionist}. This connectionist approach was largely abandoned in favour of more manageable coupled limit-cycle oscillators, where each parameter's effect on the overall behaviour is predictable \cite{ijspeert2008central,sun2021comparative,chen2023research}, particularly if the CPG outputs are mapped to workspace trajectories \cite{liu2011cpg}. Hence, the problem of gait adaptation has become a matter of designing appropriate feedbacks or learning schemes for the designated control parameters. These continuous learning approaches have been highly successful for learning stable locomotion and adapting to unseen physical environments \cite{nakamura2007reinforcement,thor2021locomotion}.

%Social adaptation requires limit cycle flexibility
Adapting to a social environment, however, is qualitatively different to adapting to a physical environment. One example of a social adaptation is imitation or learning by demonstration. This is now common in, for example, compliant robotic arms, where there are few constraints on movement \cite{ravichandar2020recent}. However, for most CPG-based legged robot controllers, the range of available limit cycles is prescribed by design. Therefore, the potential for imitation and synchronization is relatively limited in most current legged robots.

%EAs
The problem of large parameter spaces that comes with more flexible neuron models can be tackled using the same method by which nature succeeded to make animals roam the earth. Evolutionary methods are a useful tool for robotics, making use of the abundance of computing power now at our disposal \cite{bongard2013evolutionary,doncieux2015evolutionary} and new algorithms for promoting diversity of designs \cite{deb2013evolutionary,mouret-iscience2020}. This has led to advances in morphology design \cite{collins-gecco2018}, modular and soft robots \cite{cheney-gecco2013,nordmoen2021map}, and generative encodings \cite{veenstra-alife2020}.

%Evolution vs adaptation
While evolution and real-time adaptation may at first glance seem like unrelated processes working at very different time-scales, there are several ways in which evolution can facilitate adaptation. Evolution can, for example, globally optimize in the high-dimensional space of connection weights in recurrent neural networks so that a lower dimensional space of inputs encompasses a wide repertoire of output patterns \cite{funahashi1993approximation,floreano2008neuroevolution}. When applied to locomotion patterns, this dimensionality reduction can therefore simplify the task of online learning of which behaviours are most suitable in which state and environment, as is done in reinforcement learning \cite{hwangbo2019learning}, or the task of Hebbian learning of connection weights for higher-level control \cite{arena2009stdp,jouaiti2018hebbian}. Importantly, reactive controllers can also be optimized in advance for susceptibility to a wide range of inputs, allowing spontaneous compliant motion in an open-loop scenario.

% Multi-objective EAs
Single-objective evolutionary algorithms were used in early work on connectionist CPGs, generally to optimize some combination of walking speed, regularity and stability measures \cite{ijspeert2001connectionist,reil2002evolution}. These measures are often involved in trade-offs, meaning that despite the ``hands-off'' nature of evolutionary algorithms, fitness functions still needed to be carefully designed to weight each measure appropriately. The advent of multi-objective algorithms \cite{deb2013evolutionary} allows these measures to be separated into their own fitness functions, so that a diverse range of controllers is generated along the Pareto front of non-dominated solutions \cite{oliveira2011multi,liu2015evolution,wang2021parameters}. Therefore, in addition to the advantages of evolutionary algorithms for highly flexible neural controllers, multi-objective evolution in particular also allows for correlational studies about their emergent properties, and for controllers to be hand-picked for different sets of capabilities after a single optimization process.

%In this paper

In this paper, we use multi-objective evolution to test the assumption that evolving CPGs for flexibility can facilitate rapid real-time gait adaptation. This contributes to bio-inspired robotics in two ways. Firstly, we demonstrate novel virtual robot quadrupeds that can entrain their locomotion to a range of rhythmic external inputs without physical coupling and without explicit feedback. This kind of automatic adaptation to social environments via audio and visual perception is common in humans, such as the tendency to synchronize when walking together \cite{chambers2019pose}. While rhythmic entrainment to social partners has been demonstrated in virtual robots \cite{miyake2009interpersonal}, this used relatively slow phase-based feedbacks in linear oscillators. Therefore, using neuromorphic CPGs can increase the naturalistic quality of multi-robot and human-robot interaction at the level of basic behaviours.

Secondly, we show how examining correlations between emergent properties of the CPGs can aid future design.  Connectionist control systems are once again being increasingly employed in robotics, and these generally require searching through a large parameter space using automated processes such as reinforcement learning or genetic algorithms that target a desired ability \cite{rudin2021cat,medvet2022impact}. Identifying statistical trends between such abilities (costs or fitness functions) and features that can be more easily manipulated prior to optimization can greatly increase the efficiency of this time-consuming process \cite{mouret2015illuminating}.

For example, previous work with disembodied CPG architectures has suggested that both gait type and the sensitivity of oscillation period to neuron bias are important factors for entrainment ability \cite{szorkovszky2022rapid}. We further test this correlation in an embodied context, where there is a simultaneous goal of upright walking. To test whether our results are dependent on the morphology of the embodied system, we use two quadrupeds that differ in limb length. Wide applicability is important for interaction between robots optimized for different environments, or those that adapt their morphology in real time \cite{nygaard2021real}.

First, using a multi-objective genetic algorithm and a fitness evaluation during which two control parameters are swept, we evolve populations of CPGs for flexible walking speed and direction. Then, for a subset of CPGs that emphasize different components of the objective function, we incrementally evolve robust filters for rhythmic (such as audio) input. We analyze the real-time entrainment performance of the robots as a function of the CPG properties, in particular the flexibility of the gait period and pattern. Finally, we discuss the implications of our results for understanding adaptive and imitative behaviour, as well as the potentials of our approach for multi-robot systems, human-robot interaction and autonomous learning.

\section{Methods}

\subsection{Neuron model}

The neural model is based on the Matsuoka neuron \cite{matsuoka1985sustained}, a simple and popular model for robotics \cite{thandiackal2021emergence,liu2011cpg,taga1991self,kimura1999realization,fukui2019autonomous}. This is a biologically motivated yet abstract two-variable model:
\begin{eqnarray}
t_0 \frac{du_i}{dt} &=& -u_i -av_i + I_i(t) \\*
t_0 \frac{dv_i}{dt} &=& -\gamma v_i + bh(u_i)
\end{eqnarray}
where $h(u)$ is a rectified linear unit: $h(u)=0$ for $u\leq0$ and $h(u)=u$ for $u> 0$. Like many biological models, there is a fast ``spiking'' variable ($u_i$) and a slow ``recovery'' variable ($v_i$).

While this model produces patterns of spiking in a simple way, its linearity leads to a poor ability to adapt oscillation frequency \cite{jouaiti2019comparative}. Importantly, unlike biological neurons, the spiking rate is insensitive to changes in tonic input \cite{matsuoka1987mechanisms}. To address this shortcoming, we add a sigmoidal ``deactivation'' function to the fast variable $u_i$
\begin{eqnarray}
t_0 \frac{du_i}{dt} &=& -u_i -aS(\kappa[u_i-u_0]) v_i + c_i + d_i I_{DC} + I_{\mathrm{AC},i}(t) \\*
t_0 \frac{dv_i}{dt} &=& -\gamma v_i + bh(u_i) \; ,
\end{eqnarray}
where $S(x)=1/(1+\exp(x))$. In addition we introduce $I_{DC}$, a control parameter modelling the global brain stem input separately from fluctuating inputs $I_{ACi}$ and a constant offset $c_i$. For a certain parameter range satisfying
\begin{equation}
c_i + d_i I_{DC} > u_0 + \frac{2}{\kappa}   \; ,
\end{equation}
this reproduces the general nullcline shape, as well as the input-dependent firing rate, of biological neuron models \cite{skinner1994mechanisms}. The fast input is modelled:
\begin{equation}
I_{\mathrm{AC},i}(t) = G_i I_{\mathrm{in}}(t) + I_{\mathrm{fb},i}(t) + \sum_{j \neq i} w_{ij}h(u_j(t) - \tau_{ij})
\end{equation}
where $I_\mathrm{fb,i}$ is sensory feedback and $I_{in}$ is the external input, $G_i$ is input sensitivity, and $w_{ij}$ is the synaptic weight and $\tau_{ij}$ the threshold for a connection from neuron $j$ to neuron $i$.

\begin{figure}[!t]
\centering
\includegraphics[width=7cm]{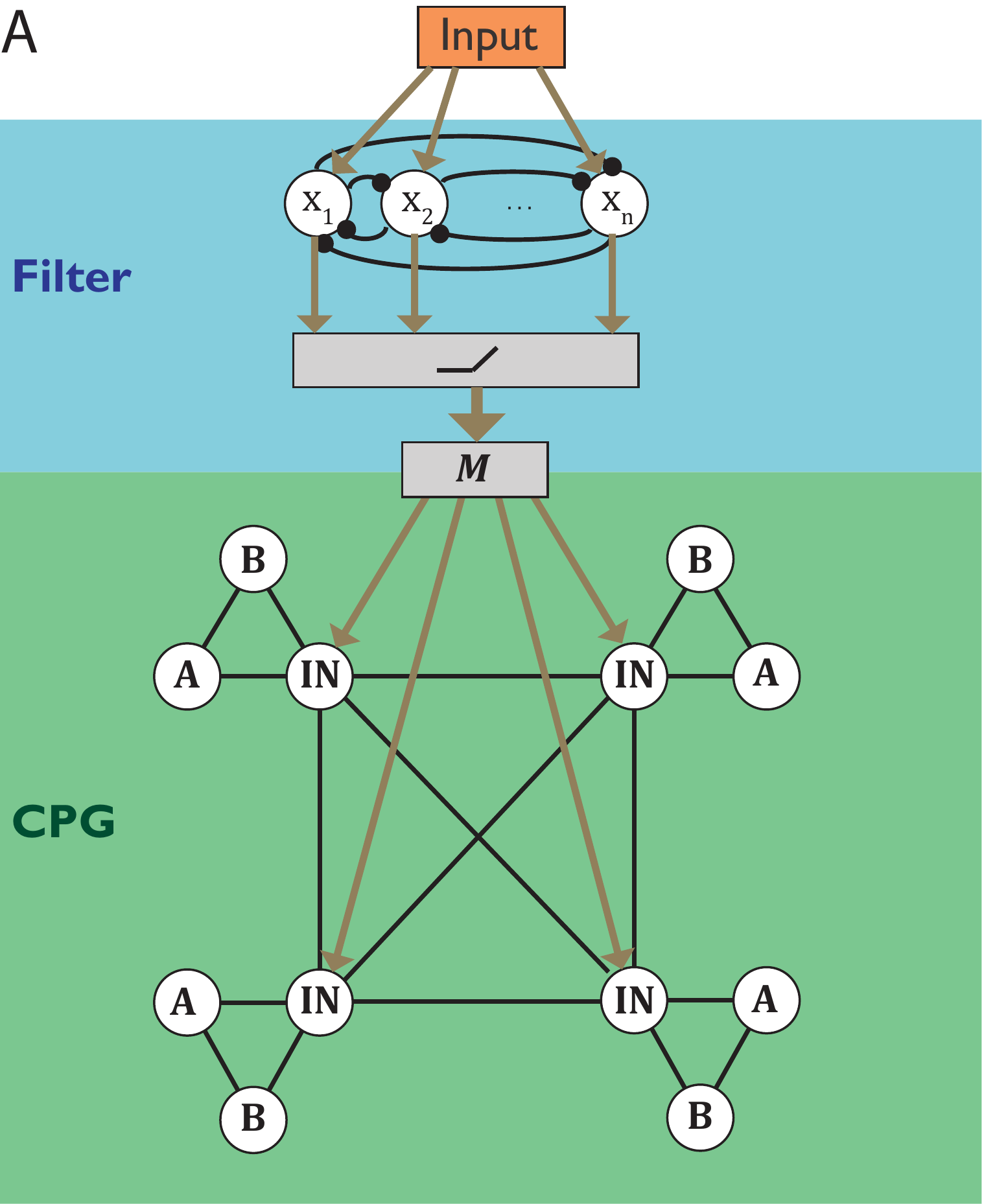}
\includegraphics[width=5cm]{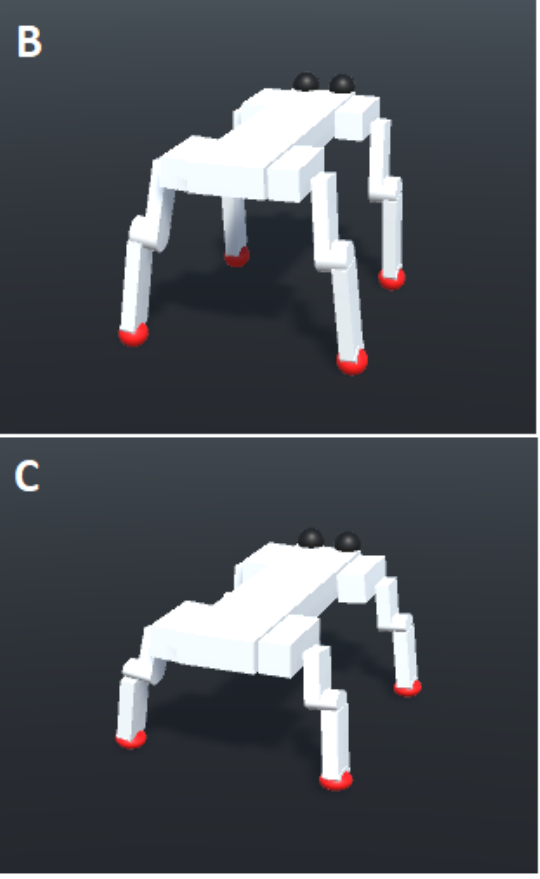}
\caption{Panel \textbf{(A)} shows the layout of the robot controller. Circles denote modified Matsuoka neurons ($n$ in the filter component, and 12 in the central pattern generator). Arrows denote one-way connections (either excitatory or inhibitory), lines ending in circles denote inhibitory connections, and regular lines denote mutual connections that may be excitatory or inhibitory. The modules are connected via $n$ rectifying linear units with threshold $\tau_0$, followed by an $n$ by 4 weight matrix $\mathbf{M}$. A/B: motor neurons; IN: interneurons. For this study, $n=6$. \textbf{(B)}: Open dynamic robot-based design. \textbf{(C)}: Short-legged variant.}\label{fig:schematic}
\end{figure}

\subsection{CPG and Filter modules}

The layout of the quadruped is shown in Figure \ref{fig:schematic}A. This is a simplified version of the CPG layout in \cite{danner2017computational} containing one flexor-extensor pair per limb and several interneuron types. Our simplified model consists of three neurons per limb: one interneuron, a leg joint neuron (A) and a knee joint neuron (B). Each limb has identical parameters, and the connection weights are constrained to obey lateral symmetry. The ranges of the 23 neuron parameters and connection weights are given in Supplementary Table S1. For the CPG module, $G_i=0$ and $\tau_{ij}=0$ within the CPG. Depending on the connection weights, which could be excitatory or inhibitory, and the brain stem drive, the CPG can autonomously generate coordinated oscillations in the motor neurons. Flexible movement was targeted for this autonomous behaviour in the first stage of evolution (see section \ref{cpgevo}).

A subset of the evolved CPGs with high fitness had filter modules evolved on top (see Section \ref{filterevo}). The purpose of this layer is to preprocess and distribute descending rhythmic signals $I_{in}(t)$ for the CPG to entrain its oscillations to. This module had only inhibitory connections, no lateral symmetry, no feedback and no brain-stem control ($d_i=0$). A non-zero threshold $\tau_{ij}=\tau_0$ was imposed between the filter outputs and CPG inputs so that the CPG received no input from the filter when $I_{in}(t)=0$. The modules are unidirectionally linked by the weight matrix $\mathbf{M}$, with a wider range than the within-module weights $w_{ij}$ (see Supplementary Table S2).

\subsection{Robot simulation}

The quadruped robots were simulated in the Unity game engine on a flat planar surface. The full-scale robot was based on the specifications of the Open Dynamic Robot \cite{grimminger2020open}. The `short-legged' version was identical, apart from a 40\% reduction in upper leg length and a 33\% reduction in lower leg length. The controllers were written in Python and interfaced with Unity using the Unity ML-agents package \cite{juliani2018unity}.

The CPG used a time interval of 8 ms while the physics simulation used a time interval of 20 ms, with decisions made every $\Delta t=$100 ms. At each decision point, the CPG sent joint positions based on the changes in rectified motor neuron outputs since the previous simulation step $\Delta h(u_A(t))$ and $\Delta h(u_B(t))$:
\begin{eqnarray}
\theta_\mathrm{leg}(t) &=& \theta_{0,\mathrm{leg}} + \theta_{\mathrm{lim,leg}} \left[2S\left(\frac{2A}{\theta_\mathrm{lim,leg}} \frac{\Delta h(u_A(t))}{\Delta t} \right)-1\right] + \theta_{C} \\
\theta_\mathrm{knee}(t) &=& \theta_{0,\mathrm{knee}} + \theta_{\mathrm{lim,knee}} \left[2S\left(\frac{2B}{\theta_\mathrm{lim,knee}} \frac{\Delta h(u_B(t))}{\Delta t} \right)-1\right]
\end{eqnarray}
where $S(x)$ is a logistic function, limiting the half-amplitude of motion of the upper leg and limb joints to $\theta_\mathrm{lim,leg}$ and $\theta_\mathrm{lim,leg}$, respectively, both of which are set to 90 degrees. The coefficients $A$ and $B$, and the zero-angles are allowed to evolve, however with leg zero-angles $\theta_{0,\mathrm{leg}}$ always positive and $\theta_{0,\mathrm{knee}}$ always negative, corresponding to a full-elbow pose. The hip joint, perpendicular to the leg and knee joints, was kept at a constant but evolvable parameter $\theta_{0,\mathrm{hip}}$. See Supplementary Table S3 for the ranges of these evolvable parameters. A control parameter $\theta_C$ was also added to the leg joint angle so that the forward position of the center of mass could be controlled in real time (see next section).

%In addition to the 23 CPG parameters, 9 additional evolvable parameters were used for the interface. These were the hip joint angle ($0.03$ to $0.3$ radians), the upper leg joint standing angle ($-0.05$ to -$0.5$  radians), the knee leg joint standing angle ($0.08$ to $0.8$ radians), the amplitudes of the upper leg and knee joints ($0.05$ to $0.5$ radians per unit neuron activation at center), and four feedback amplitudes  ($-0.45$ to $0.45$).

At each decision point, the CPG also received inputs processed from the body's tilt, to use as stabilizing feedback inputs $I_\mathrm{fb,i}$.  The sideways tilt (the sideways component of the unit vector normal to the top of the body) was input with opposite signs to the left and right limb motor neurons, with separate coefficients $q_\mathrm{A,side}$ and $q_\mathrm{B,side}$ for neurons A and B, respectively. Likewise, the front-back tilt (the upwards component of the unit vector normal to the front of the body) was input with opposite signs to the front and back limb inputs, with coefficients $q_\mathrm{A,front}$ and $q_\mathrm{B,front}$. These four coefficients were also evolved along with the CPG.  The entire set of 32 parameters for the CPG and body was encoded as a sequence of integers, each taking a value between 1 and 10.

\begin{figure}[!t]
\begin{center}
\includegraphics[width=7.5cm]{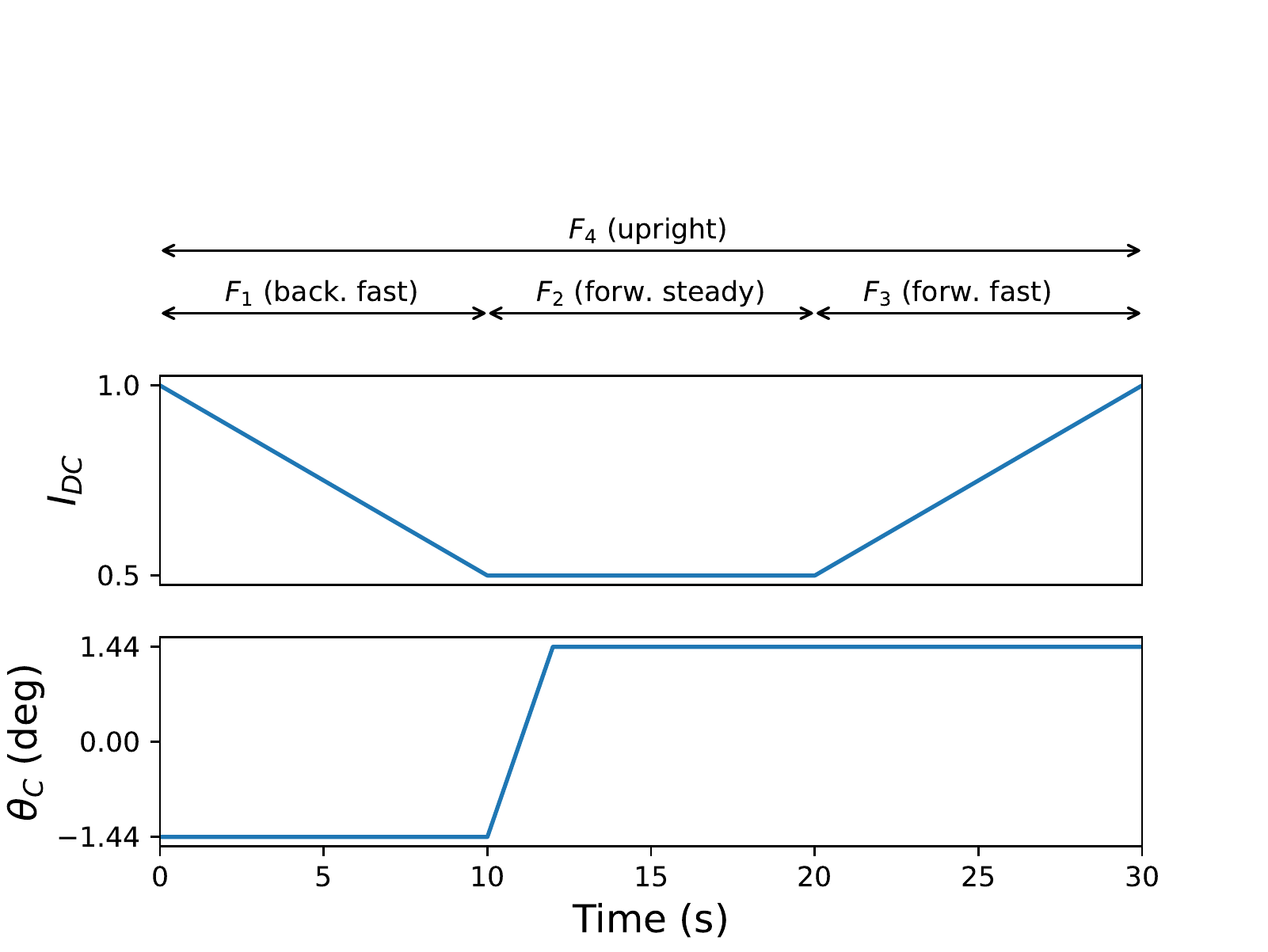}
\includegraphics[width=6.5cm]{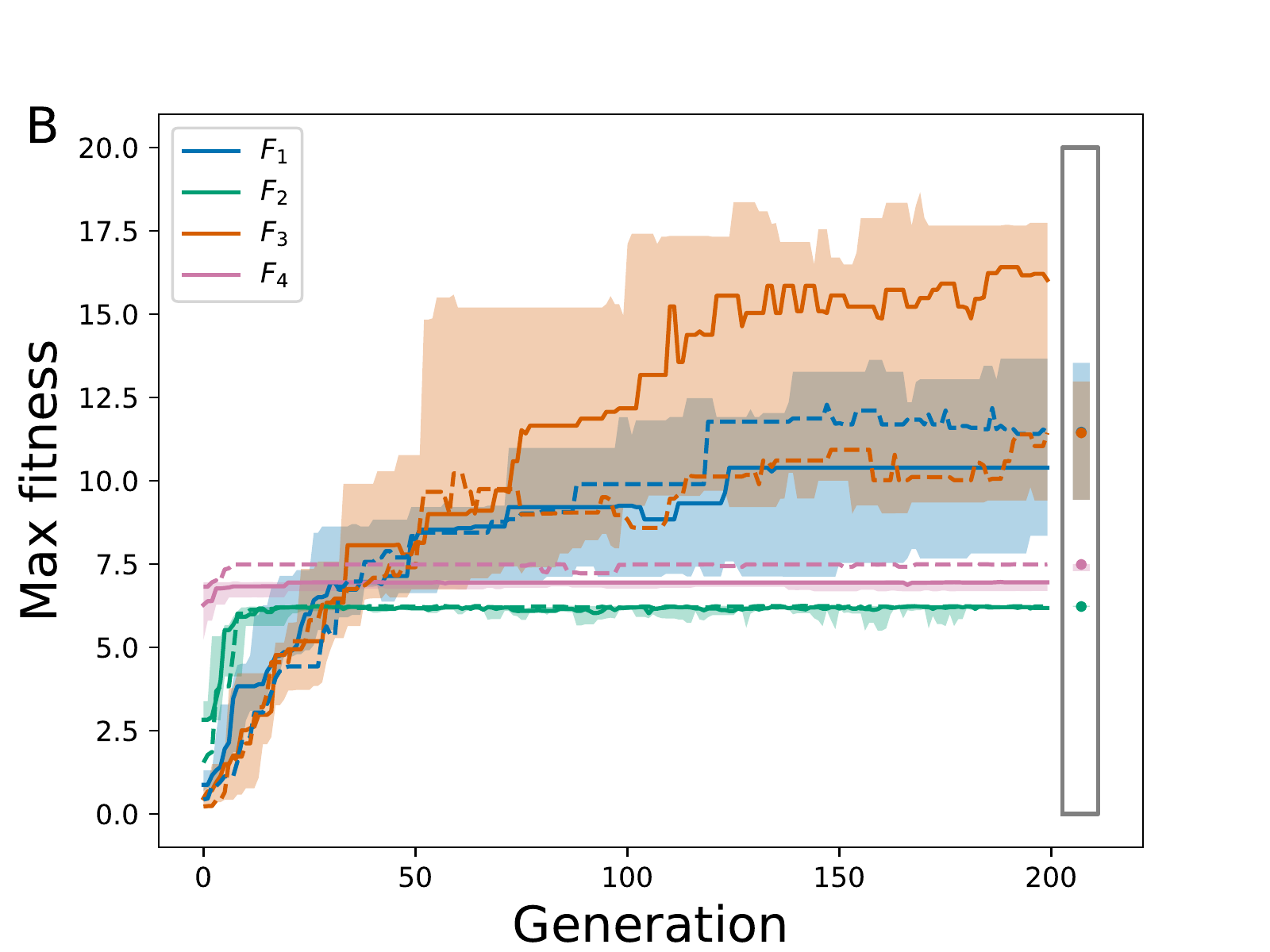}
\end{center}
\caption{CPG evaluation and evolution. \textbf{(A)} Sweeping of control parameters $I_{DC}$ and $\theta_C$ over the course of a single evaluation. The CPG undergoes a `burn-in' period of 8s prior to the $t=0$ mark. For the first two seconds of the simulation, the actuators are ramped from zero to the full CPG output. The arrows above show the measured fitnesses and targeted behaviours during each stage. \textbf{(B)} Maximum fitnesses vs generation. Solid lines: normal quadruped; dotted lines: short-legged variant. Each line is the median over 5 replicates. Shaded areas are the range over the 5 replicates for the normal quadruped. Areas inside the box on the right side of the plot are the final ranges of maximum fitness for the short-legged variant.}\label{fig:evolution}
\end{figure}

\subsection{CPG evolution}
\label{cpgevo}
We used the NSGA3 algorithm \cite{deb2013evolutionary} in the DEAP Python package \cite{DEAP_JMLR2012} to perform multi-objective optimization. This genetic algorithm preferentially selects non-dominated individuals (i.e.\ those on the Pareto front) for propagation to the next generation. Parameters used for the NSGA3 algorithm are given in Supplementary Table S4.

Unlike in \cite{szorkovszky2022rapid} where each CPG was iteratively evaluated to explicitly select for change in period as a function of the brain stem drive $I_{DC}$, we instead swept $I_{DC}$ during a single evaluation, and partially selected for variability in speed. This was to reduce the computational load of running several evaluations with constant parameters as is required for reliable period estimation, and to ensure stable walking over a wide region of parameter space. In addition, we adjusted the center of mass parameter $\theta_C$ that is added to the leg standing angle during the evaluation, in order to select for flexibility of movement direction. Measurements of forward and perpendicular distance covered ($y_j$ and $x_j$ respectively) were made over three stages ($j=1$ to $j=3$) of length 10 seconds each, as shown in Figure \ref{fig:evolution}A.

We used four fitness functions to simultaneously select for desired capabilities of the robot. The first three correspond to the targeted behaviours for each stage of the evaluation (fast backward motion, steady forward motion and fast forward motion, respectively), while the fourth selects for overall stability. The fitnesses are given as:
\begin{eqnarray}
F_1 &=& -y_1 - \left(\frac{x_1}{x_0}\right)^2 \label{eq:F1} \\*
F_2 &=& 2y_0 y_2 - y_2^2 - \left(\frac{x_1}{x_0}\right)^2 \label{eq:F2}\\*
F_3 &=& y_3 - \left(\frac{x_1}{x_0}\right)^2 \label{eq:F3} \\*
F_4 &=& \frac{y_0^2 H_\mathrm{tot}}{1 + t_\mathrm{tot}} \label{eq:F4}
\end{eqnarray}
where $x_0=\sqrt{5}$ m is a parameter to punish sideways movement, $y_0=2.5$ m is an optimal forward distance for steady motion, defining the maximum $F_2$ and $F_4$, $H_\mathrm{tot}$ is the mean height over the entire evaluation (normalized for a maximum of $\approx 1$), and $t_\mathrm{tot}$ is the root mean square body tilt (mean length of the cross product $\hat{\bm{n}} \times \hat{\bm{g}}$ where $\hat{\bm{n}}$ is the unit vector normal to the top of the robot and $\hat{\bm{g}}$ is the unit vector normal to the ground).

Note that $F_1$ and $F_3$ are unbounded and identical but with opposite signs for the forward distance. $F_2$, meanwhile, is a quadratic function that is positive only between $y_2=0$ and $y_2=2y_0$. To optimize all four fitness functions simultaneously, the robot must first walk backwards with a negative center of mass parameter, then walk forwards $2.5$ m with a positive center of mass parameter and brain stem input of $I_{DC} = 0.5$, and then accelerate with $I_{DC}$ increasing.

During the evolution, the evaluation was run three times per individual with random initial $u_i$ values, and the median of each fitness was taken. For each morphology, five independent populations of 168 individuals (8 individuals per edge of the reference Pareto front) were evolved for 200 generations. The final populations were then evaluated 15 times with different random number generator seeds and the median fitnesses were calculated again, as well as cross-correlations between limbs, and oscillation periods from the largest autocorrelation peak (see Supplementary Material).

\subsection{Filter evolution}
\label{filterevo}
In order to reduce the total number of evolutions for the filter layer, a subset of CPGs was chosen from each population's Pareto front in order to capture a numerically small but diverse range of solutions. Each population was first reduced to a set of CPGs for which all fitnesses were positive, and then four were chosen from each using the maxima of four weighted fitness functions $F_m^*$. These are defined as a combination of $F_m$ and the total sum of fitnesses:
\begin{equation} \label{maxfit}
F_m^* = zF_m + \sum_{k=1}^4 F_k \; ,
\end{equation}
where $z$ was incremented in intervals of one until the maximum of each $F_m^*$ was unique.

For each of these CPGs, a 6-neuron, a filter module was evolved using the NSGA3 algorithm. The parameters consisted of 6 input weights, 24 output weights, 30 inhibitory connection weights within the layer, a shared bias term $c_i$, and the time constant of a low-pass filter for the initial input.

The filter evolution comprised two stages of increasing complexity. The input consisted of evenly spaced impulses with every fourth impulse missing, over a total of 40 seconds. For the first 50 generations, the timings had no noise, in order to facilitate the random generation of suitable filters. After the 50th generation, a random timing offset (Gaussian distributed with a standard deviation of 2\% of the period) was applied to each impulse's timing. Evaluations were made with constant control parameters $\theta_C=0$ and $I_{DC}=0.5$.

Three input periods were used for each evaluation: $T_0/\phi$, $T_0$ and $\phi T_0$, where $T_0$ is the CPG period at $\theta_C=0$ and $I_{DC}=0.5$, and $\phi=0.618$. The latter was chosen so that $1/\phi \approx 1+\phi$ and hence the low-period and high-period inputs are equidistant from an integer multiple of $T_0$.

The fitness function for each input period $T_\mathrm{in,k}$, with measured walking period $T_\mathrm{out,k}$, was calculated as
\begin{equation}
F_{fk} = H_\mathrm{tot,k} Q_k
\end{equation}
where
\begin{equation}
Q_k = \left(1 + \frac{1}{\epsilon} \left|\frac{2T_\mathrm{out,k}}{T_\mathrm{in,k}}-\left[\frac{2T_\mathrm{out,k}}{T_\mathrm{in,k}}\right]\right| + \frac{\sigma_0}{\sigma_t}\right)^{-1} \; ,
\end{equation}
$[.]$ indicates rounding to the nearest integer, $\sigma_0$ is the mean standard deviation of the filter output with no input, and $\sigma_t$ and $\epsilon$ are scaling thresholds, both set to $0.1$ for the current study. Hence, the fitness is maximized for upright walking with a period of a half-integer or integer multiple of the input period.

The filter evolution used a population of 92 individuals (12 individuals per edge of the reference Pareto front) for 150 generations. At the final generation, the population was evaluated 5 times and the median fitnesses were calculated. This final evaluation included two additional periods at $T_0/\sqrt{\phi}$ and $\sqrt{\phi} T_0$. The filter with the highest $\sum_k Q_k$ was then chosen for each CPG, conditional on each $H_\mathrm{tot,k}$ being above $0.75$.

\section{Results}

\subsection{CPG evolution}

From the CPG evolution, 710 unique individuals were produced. As shown in Figure \ref{fig:evolution}B, the fitnesses with upper limits ($F_2$ and $F_4$) reached these within a few generations, while the others reached a plateau close to the 200 generation mark.

\subsection{Gait characteristics}

After filtering out CPGs with an average height of $< 0.75$ (below which robots were typically judged to be crawling rather than walking, see Supplementary Figure S1), CPGs were classified into gait types. Walking gaits were defined as those with maximum inter-limb correlation $<0.3$; above this threshold, trotting gaits were defined as those with diagonally opposite limbs maximally correlated, pacing gaits had left or right leg pairs maximally correlated, and bounding gaits were defined as those with front or back limbs maximally correlated. Of the 420 non-crawling individuals at $I_{DC}=0.5,\theta_C=0.016$, $19.8$\% were classed as walking, $73.8$\% as trotting, $2.4$\% as pacing and $4.0$\% as bounding. The longer legged robot was more likely to develop a walking gait (29\% vs 6\%) or bound gait (7\% vs 0\%), while the short-legged variant was more likely to develop a pacing gait (6\% vs 0\%).

\begin{figure}[!t]
\begin{center}
\includegraphics[width=7cm]{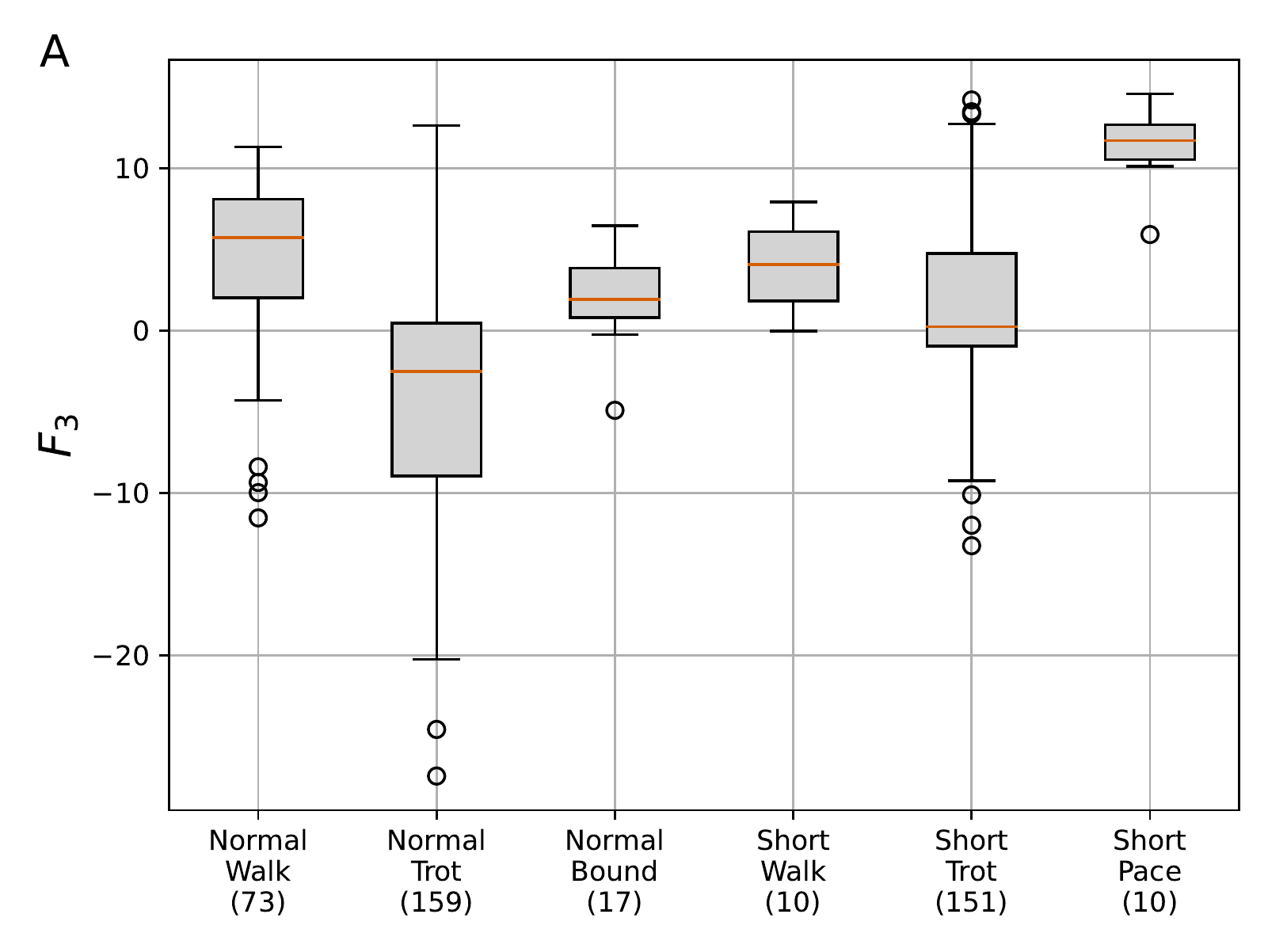}
\includegraphics[width=7cm]{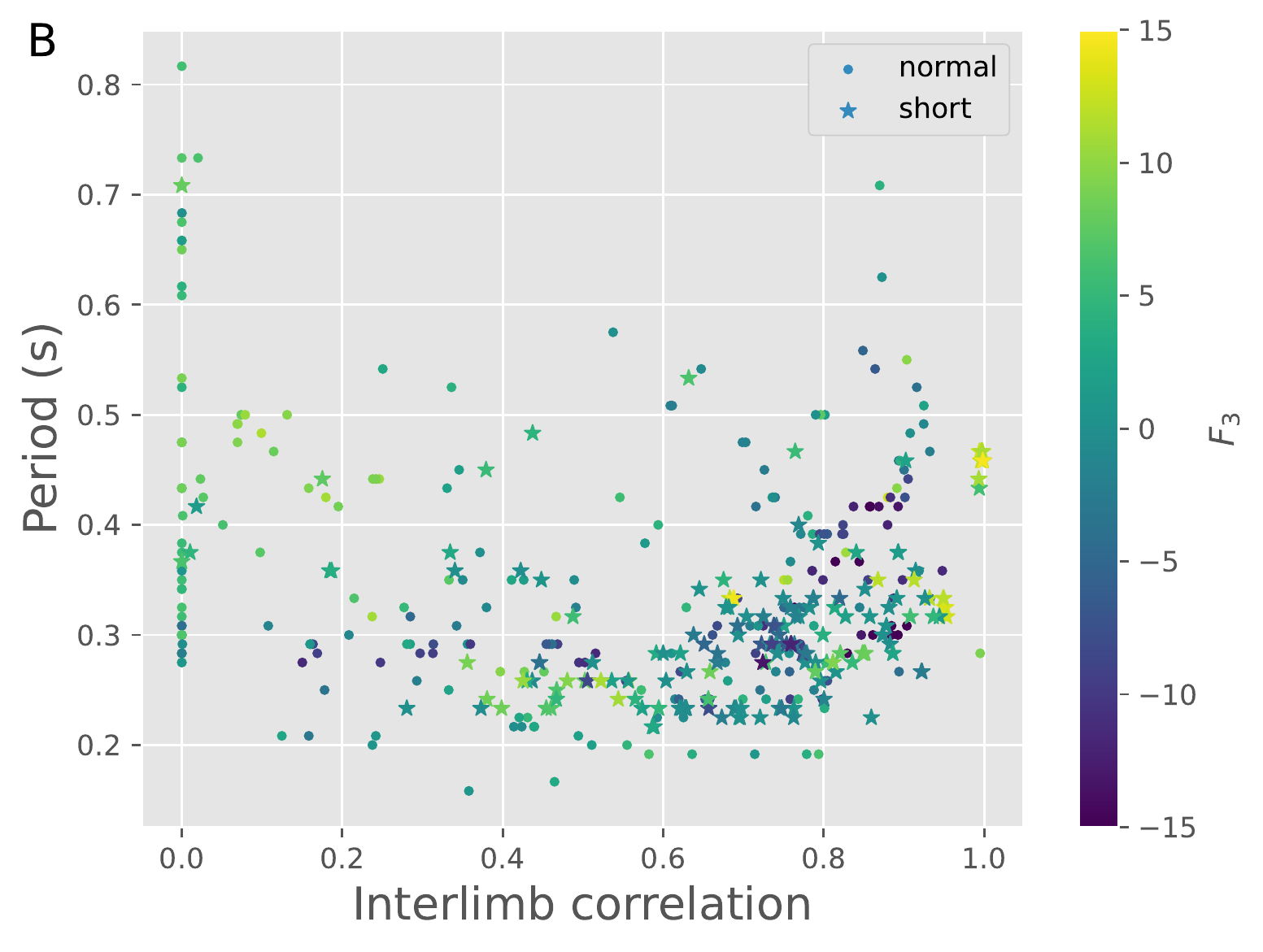}\\
\includegraphics[width=13.5cm]{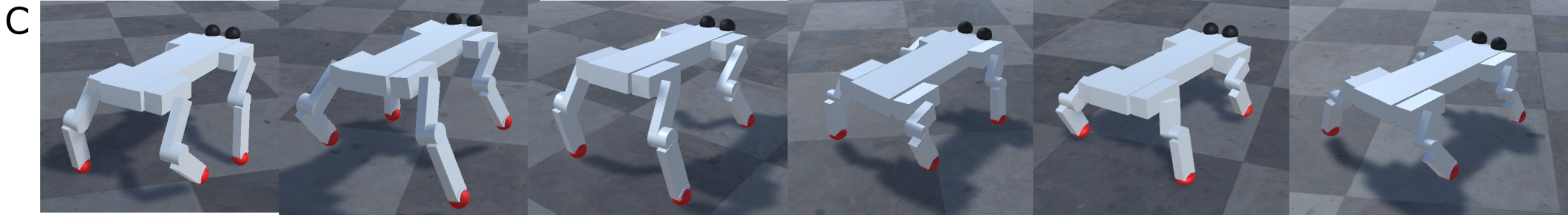}
\end{center}
\caption{$F_3$ as a function of morphology and CPG properties, for all upright robots. The box plot \textbf{(A)} shows the distributions of $F_3$ sorted by morphology and gait types. Numbers in parentheses indicate the number of each gait type in the population. In panel \textbf{(B)} $F_3$ is plotted as a marker colour against the oscillation period and maximum interlimb correlation of each CPG. For both plots, CPGs are evaluated at $I_{DC}=0.5, \theta_C=0.016$. Panel \textbf{(C)} shows images from the simulation for example individuals displaying each of the gaits and body combinations in panel A.\label{fig:F3}}
\end{figure}

\subsection{Predictors of F3}

Many CPGs (14\%) were able to have all-positive fitnesses, which means that they could walk backwards and then forwards at a controlled speed as the leg standing angle $\theta_C$ is switched from negative to positive. Of particular interest for entrainment is $F_3$, the target for acceleration with increasing $I_{DC}$. This was found to be significantly larger on average in walking and pacing gaits compared to trotting and bounding, contrary to the typical order of quadruped gaits (see Figure \ref{fig:F3}A).

\begin{figure}[!t]
\begin{center}
\includegraphics[width=15cm]{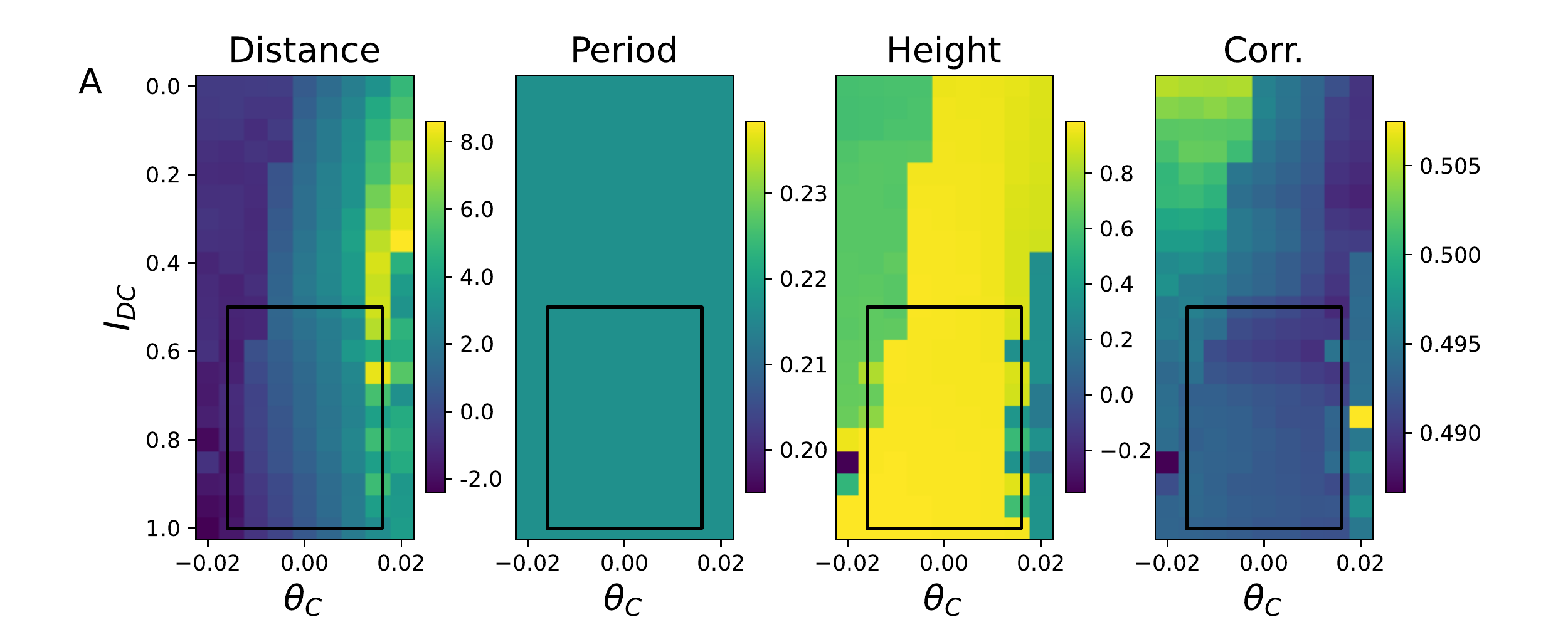}
\includegraphics[width=15cm]{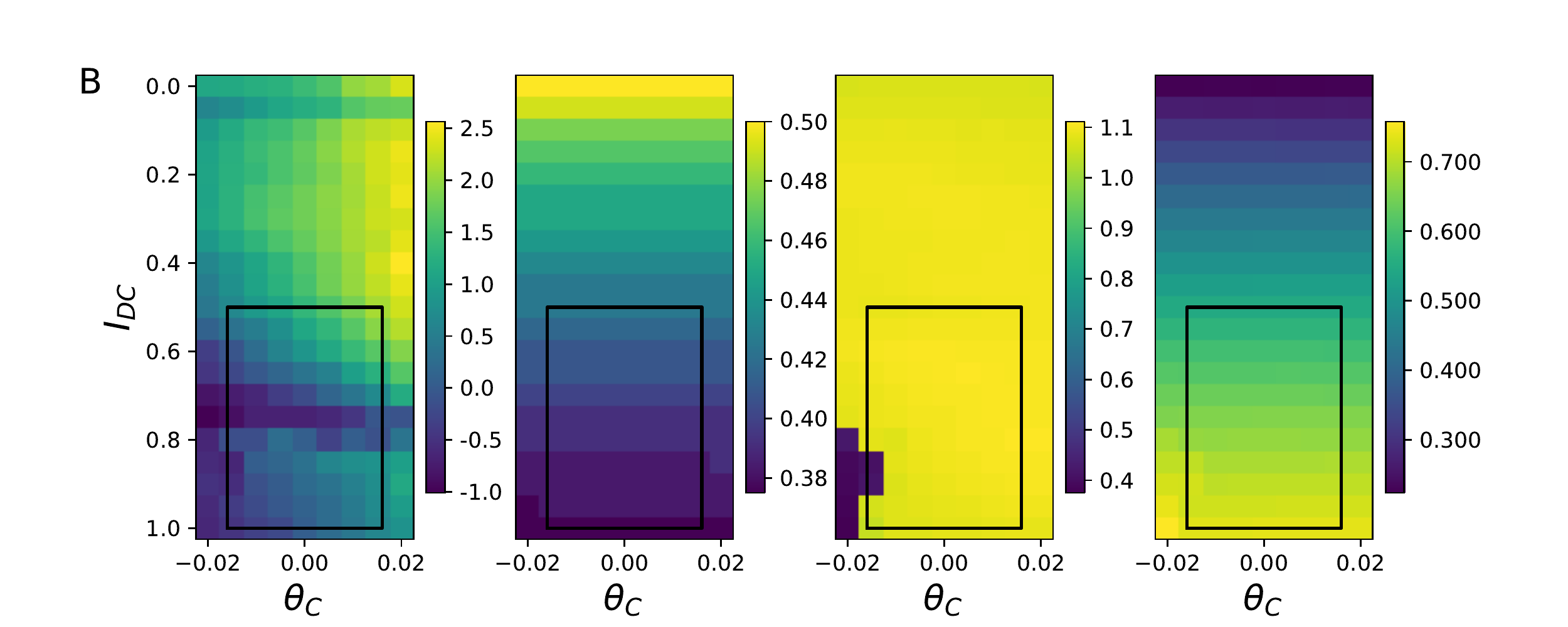}
\end{center}
\caption{Properties of the CPG with \textbf{(A)} the highest average fitness and \textbf{(B)} the highest negative change in period with $I_{DC}$ as a function of the two control parameters when both are held constant. The black line indicates the range of parameters swept during the evolution. Corr.: Maximum inter-limb correlation coefficient.}\label{fig:heatmaps}
\end{figure}

We examined whether $F_3$ selected for period or gait flexibility as predicted. For both morphologies, period and $F_3$ had a non-monotonic relationship with interlimb correlation at $I_{DC}=0.5$, as shown by Figure \ref{fig:F3}B. A clear correlation was seen, however, between $F_3$ and the change in a CPG's inter-limb correlation as the brain-stem drive was changed from $I_{DC}=0.5$ to $I_{DC}=1.0$. This was shown by a linear mixed-effect model with sums and differences of the periods and inter-limb correlations as fixed effects, and replicate as a random effect (long-legged: $z=-2.17, P=0.03$; short-legged: $z=-3.26, P=0.001$, see Supplementary Figure S3). This indicates that acceleration leading to a high $F_3$ could be achieved by moving from a correlated trotting gait towards a more efficient walk-like gait. In addition, for the shorter legged morphology, a period that shortens with increasing $I_{DC}$ is associated with higher $F_3$ score ($z=-2.72, P=0.007$, see Supplementary Material).

\begin{figure}[!t]
\begin{center}
\includegraphics[width=7cm]{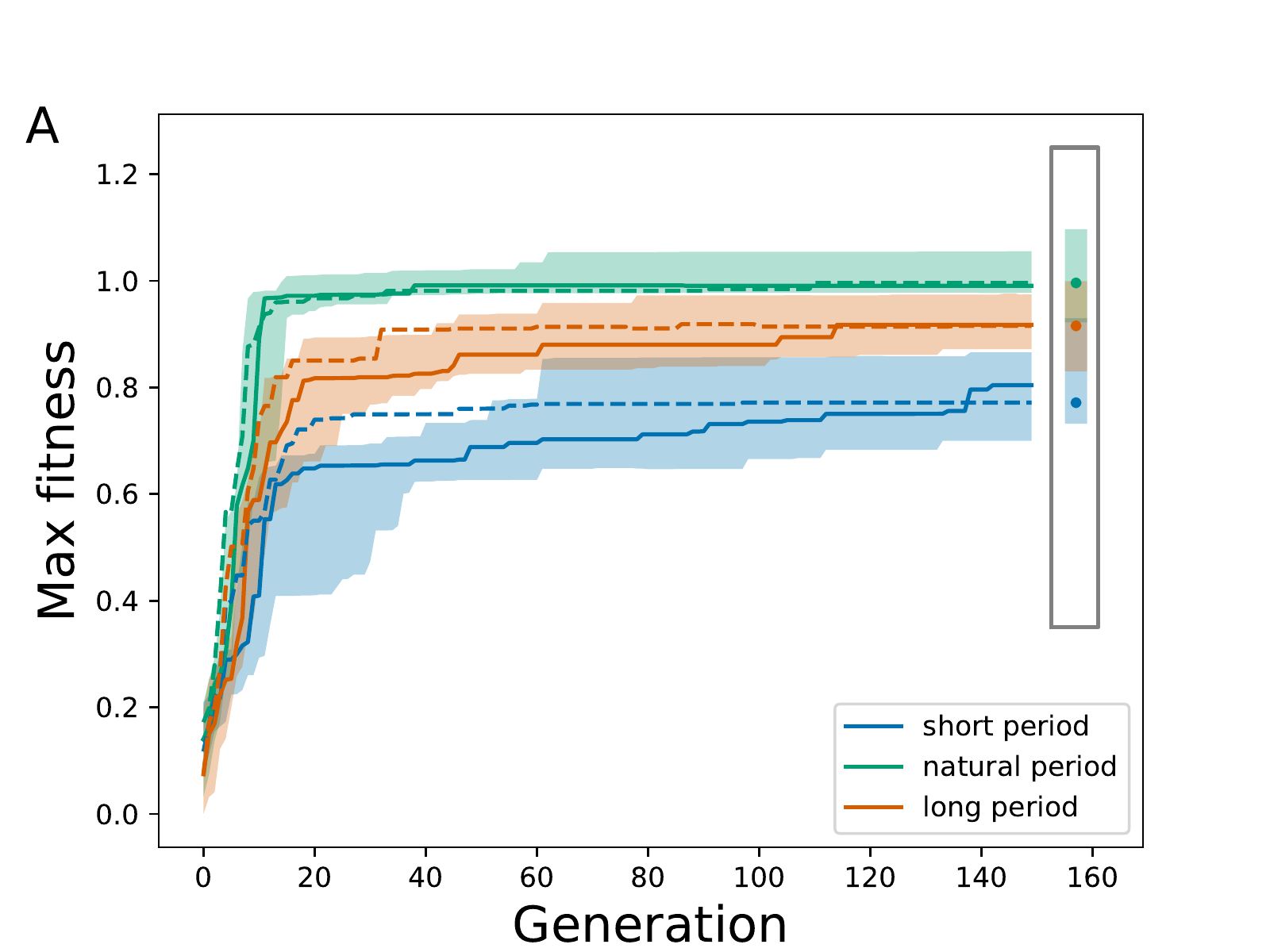}
\includegraphics[width=7cm]{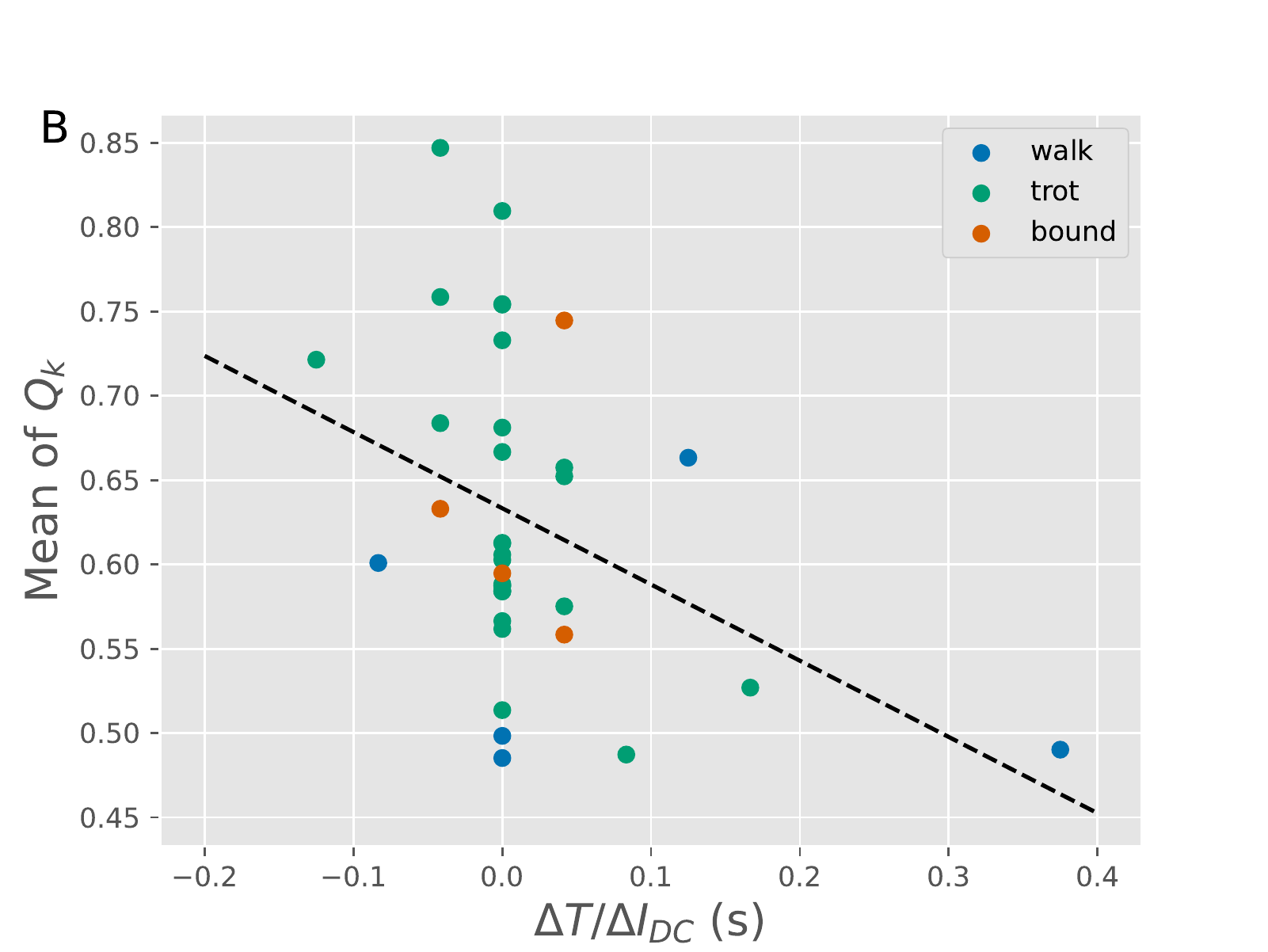}
\end{center}
\caption{Filter evolution. Panel \textbf{(A)} shows the maximum fitnesses $F_{fk}$ vs generation for the three input periods. Solid lines: normal quadruped; dotted lines: short-legged variant. Each line is the median over 19 CPGs used as the basis. Shaded areas are the interquartile range over the 19 CPGs for the normal quadruped. Areas inside the box on the right side of the plot are the final interquartile ranges of maximum fitness for the short-legged variant. Panel \textbf{(B)} shows the total entrainment ability score $\sum_{k=1}^3 Q_k/3$ for all robots over the height threshold for each $k$, as a function of period flexibility (gradient in period vs brain stem drive, with $\Delta I_{DC} = 0.2$). The dotted line is the best fit from the linear regression.}\label{fig:filter}
\end{figure}

\subsection{Direction and speed tuning}

The trained robots typically show regions of smooth change of speed and direction within the space of control parameters. These regions often extend outside the region of the control parameter sweeps. Two examples are shown in Figure \ref{fig:heatmaps}. For the CPG that maximized the average CPG evolution fitnesses (Eqs \ref{eq:F1}-\ref{eq:F4}), the brain stem drive $I_{DC}$ has relatively little effect on the movement characteristics. This illustrates that high average fitness is itself not a guarantee of general flexibility. Further outside the region of the swept control parameters, the behaviour becomes more unpredictable. For example, the low measured height in Figure \ref{fig:heatmaps}A for low drive parameter and negative $\theta_C$ indicates an instability that coincides with a gait transition, as shown by an abrupt change in inter-limb correlation in the same region.

\subsection{Filter evolution}

Each morphology had 19 CPGs chosen for filter evolution out of the planned 20. One long-legged replicate only produced three unique CPGs from Equation \ref{maxfit}, while one selected CPG from the short-legged replicates had no measurable walking period, so an input period could not be determined.

The evolution of the filter module is shown in Figure \ref{fig:filter}A. The short-legged morphology converged to the maximum fitness sooner than the long-legged morphology. In general, it was more difficult to entrain to a rhythmic input shorter than the natural walking period, compared to a longer period input.

\subsection{Predictors of entrainment}

\begin{figure}[!t]
\begin{center}
\includegraphics[width=11cm]{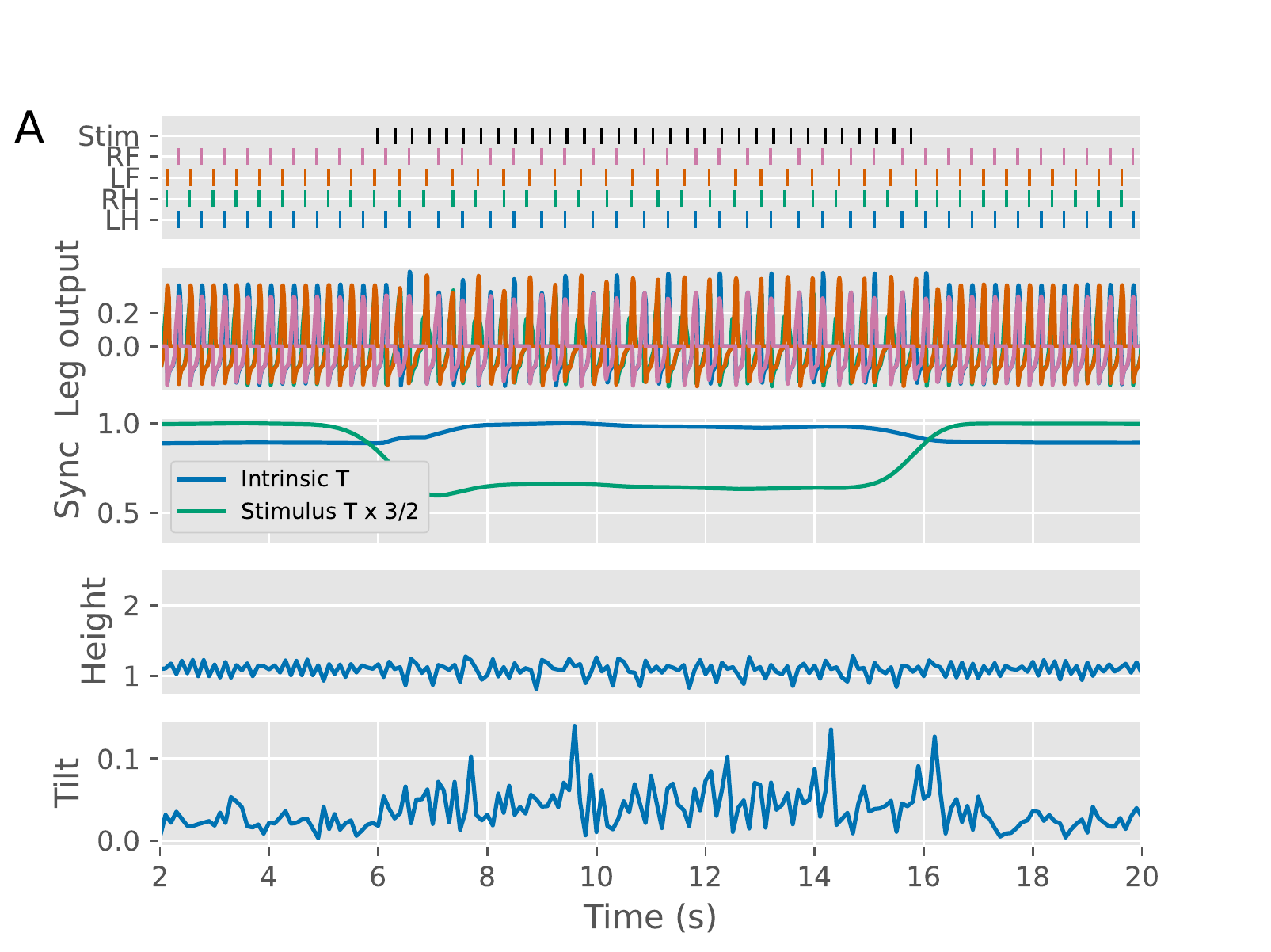}
\includegraphics[width=11cm]{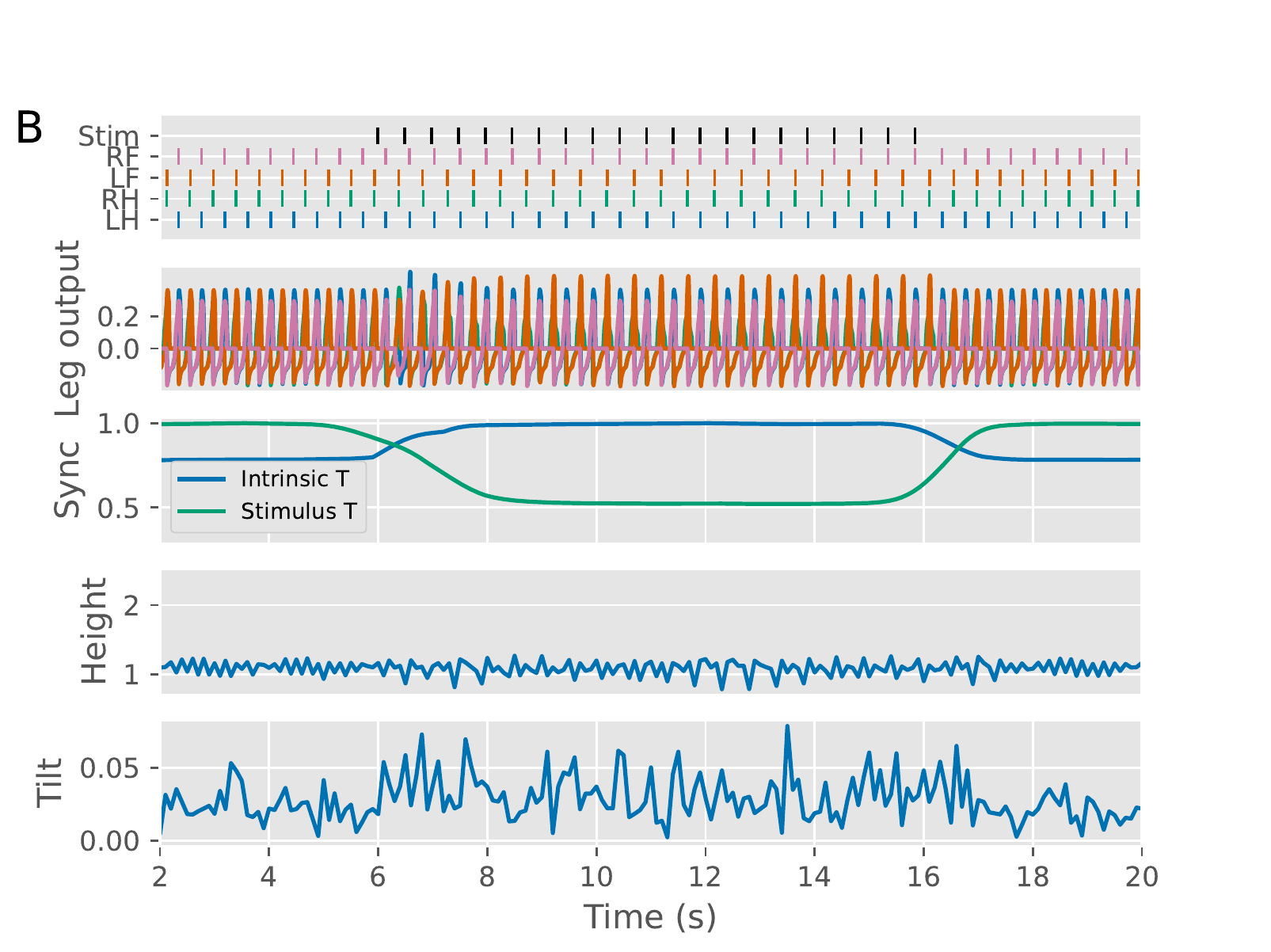}
\end{center}
\caption{Entrainment to an isochronous stimulus at \textbf{(A)} 80\% and \textbf{(B)} 125\% of the natural period, respectively, for the CPG with highest negative change in period with $I_{DC}$. The stimulus is started at the 8 second mark, and ends at the 16 second mark. Black ticks show the impulse times for the stimulus, and the peaks of the leg output below, corresponding to the extent of the forward swings in radians. Sync: output from the smoothed Morlet wavelet convolution, normalized to a maximum of one.} \label{fig:timeseries}
\end{figure}

When taking the highest eligible mean of the entrainment performance $Q_k$, a correlation was found between the period tunability and entrainment performance (linear model: $t=-2.26,P=0.03$, see Figure \ref{fig:filter}B). Faster oscillation with increasing $I_{DC}$ therefore facilitates entrainment, while faster oscillation with decreasing $I_{DC}$ appears to inhibit this ability. For the highest fitness filter and CPG combinations, entrainment could be generalized to stimulus periods other than those used during evolution, as shown in Figure \ref{fig:timeseries}. Adjustment to the stimulus turning on or off typically occurred within a few motion cycles. To show this, time series for the leg joint angles were convolved with a Morlet wavelet at the input period, with a resolution parameter $\sigma=1.5$, and then a Gaussian filter was applied with a width of $0.5$ s. Interestingly, the robot could generate a response that created a polyrhythm with the input (namely, two steps for every three impulses), as shown in Figure \ref{fig:timeseries}A.

\section{Discussion}

Our highly nonlinear, bio-inspired central pattern generator combined with multi-objective optimization was successful in both generating a variety of gait profiles and properties, as well as flexibility in these gaits for a large subset of individuals. This was despite the fact that the fitness functions did not straightforwardly translate to specific gait properties. 

Trots were the most favoured gait type for both morphologies, seemingly due to this gait's ability to transition from backwards to forwards motion. The short-legged robot was more predictable, as evidenced by its smaller range of fitnesses for each gait type, and evolved faster in the case of the filter layer. Hence, it can be used as a starting point for building complex behaviour in stages \cite{bongard2011morphological}.

Notably, the gait frequency emerged in a self-organized fashion from the interactions in the CPG network and --- due to the added nonlinearities of the neural model --- was also sensitive to inputs. Oscillation periods could therefore be tuned both manually via the brainstem drive parameter, and automatically via spontaneous entrainment to fluctuating sensory input. Due to the fully self-organized nature of the entrainment, this occurred much more rapidly than feedback-based approaches \cite{buchli2006finding,brambilla2006adaptive,miyake2009interpersonal}. We expect similar results if utilizing other neuron models with input-dependent frequency, such as the Fitzhugh-Nagumo model \cite{nagumo1962active} or the Rowat-Selverston model \cite{jouaiti2018hebbian}. Phase synchronization to stimulus may be achieved in the future by adding feedback from reaction forces, such as with the {\em Tegotae} approach \cite{owaki2017minimal}.

The specific results in this article can also be used to further direct the evolution of desired behaviours. For example, entrainment appears to benefit from a period that decreases with input, as occurs in single biological neurons. Restricting parameters to ensure this can increase the speed of evolution and the likelihood of entrainment. While the fitness targeting fast forward movement was significantly correlated with period flexibility, the period ranges exhibited were not as wide as when directly using the latter explicitly as a fitness function, as was done in \cite{szorkovszky2022rapid}. Therefore, a combined approach where a disembodied CPG is first evolved may be beneficial in future work.

Rhythmic entrainment is a complex behaviour seen only in very few species \cite{bouwer2021rhythmic}, and is thought to be an integral part of the evolution of human social behaviour. Our CPG network mediated by a filter layer successfully captures this important example of cortical shaping of cyclical movements. Our heavily bio-inspired approach offers a path towards testing theories of human cognitive processes, such as beat perception, that are still not well understood \cite{koelsch2019predictive,lakatos2019new}. The results we present show that dynamic attending theory, based on synchronization of endogenous rhythms \cite{large1999dynamics, palmer2022we}, is a viable explanation for beat perception also when involving an entire distributed sensorimotor system.

On the engineering side, this approach is highly relevant to the current push towards adaptive robot behaviour \cite{aoi2017adaptive,nocentini2019survey,ravichandar2020recent}. In this study, the filter layer was optimized by a genetic algorithm. However, by instead adding Hebbian plasticity \cite{kempter1999hebbian}, reinforcement learning \cite{hwangbo2019learning} or another mechanism for longer-term adaptation, it may be possible for robots to learn suitable new movement patterns through repeated imitation of robot or human demonstrators in an unsupervised manner, and hence develop useful behaviours autonomously \cite{winfield2011embodied}.

Future work will implement physical robots with performance metrics addressing movement efficiency and stability in different physical environments. We will also explore mutual adaptation of multiple robots by using foot sensors to transmit impulses to neighbours. Notably, the fact that the communication medium is a simple time series means that co-ordination can occur across differing morphologies. This is also likely to increase the complexity of the behaviours, as is typical in studies of collective motion \cite{sumpter2010collective}, and may allow for creative uses, such as human-robot musical ensembles \cite{krzyzaniak2021musical}.

\section*{Acknowledgments}
This project has received funding from the European Union’s Horizon 2020 research and innovation programme under the Marie Sklodowska-Curie grant agreement No 101030688, and is partially supported by the Research Council of Norway through its Centres of Excellence scheme, project number 262762. The authors would also like to thank Caroline Palmer and Anne Danielsen for helpful discussions.

\section*{Data Availability Statement}
The code used and datasets generated for this study can be found at the project Github: \\
\url{https://github.com/aszorko/COROBOREES/tree/Paper2}

\section*{References}

%\bibliographystyle{IEEEtran}
%\bibliography{main}

\newpage

\renewcommand\thefigure{S\arabic{figure}}    
\setcounter{figure}{0}  
\renewcommand\thetable{S\arabic{table}}    
\setcounter{table}{0}

\section*{Supplementary Material}

\subsection*{Parameter ranges} 

Table \ref{paramtable} shows both fixed parameters (single numbers) and evolvable ranges (inside square brackets) for the CPG module. Connection weights $w_{ij}$ are zero between limbs, apart from when $i$ and $j$ are both interneurons (see main text Figure 1A).

Parameter ranges for the filter module are shown similarly in Table \ref{filtparamtable}. Here, $\Gamma$ is the time constant of an exponential low-pass filter for the impulse stimulus. 

Finally, parameter ranges for the joint activation and sensory feedback are given in Table \ref{quadtable}. These are evolved in the same genotype as the CPG parameters.

\subsection*{Evolution parameters}

NSGA-III parameters are shown in Table \ref{evoparams}. Here, $p_c$ is the crossover probability and $p_m$ is the mutation probability per allele.

\subsection*{Period and interlimb correlation}

The period was determined using the autocorrelation of a complex-valued time series $z(t) = x(t)+iy(t)$ where $x(t)$ and $y(t)$ are the leg and knee neuron outputs before the sigmoidal transform respectively, and where $t$ begins halfway through the evaluation. The time lag for the maximum correlation was extracted, with a minimum of $t_0\approx 0.05$ seconds. During the filter evolution, a maximum lag of $2.25$ times the natural period was imposed. If no peak was found in this region, the CPG was excluded from further analysis.

\begin{figure}[t]
\begin{center}
\includegraphics[width=8.5cm]{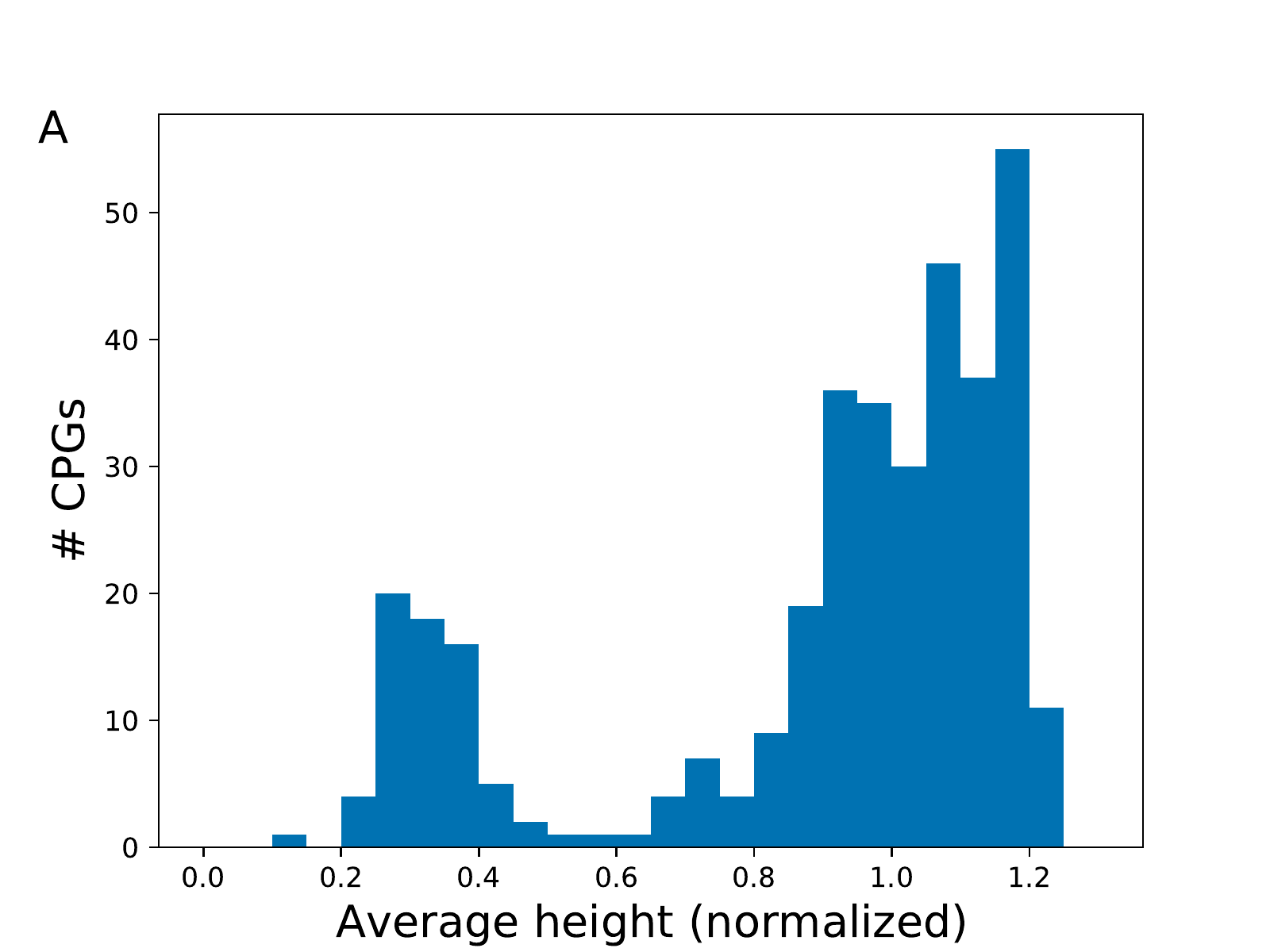}
\includegraphics[width=8.5cm]{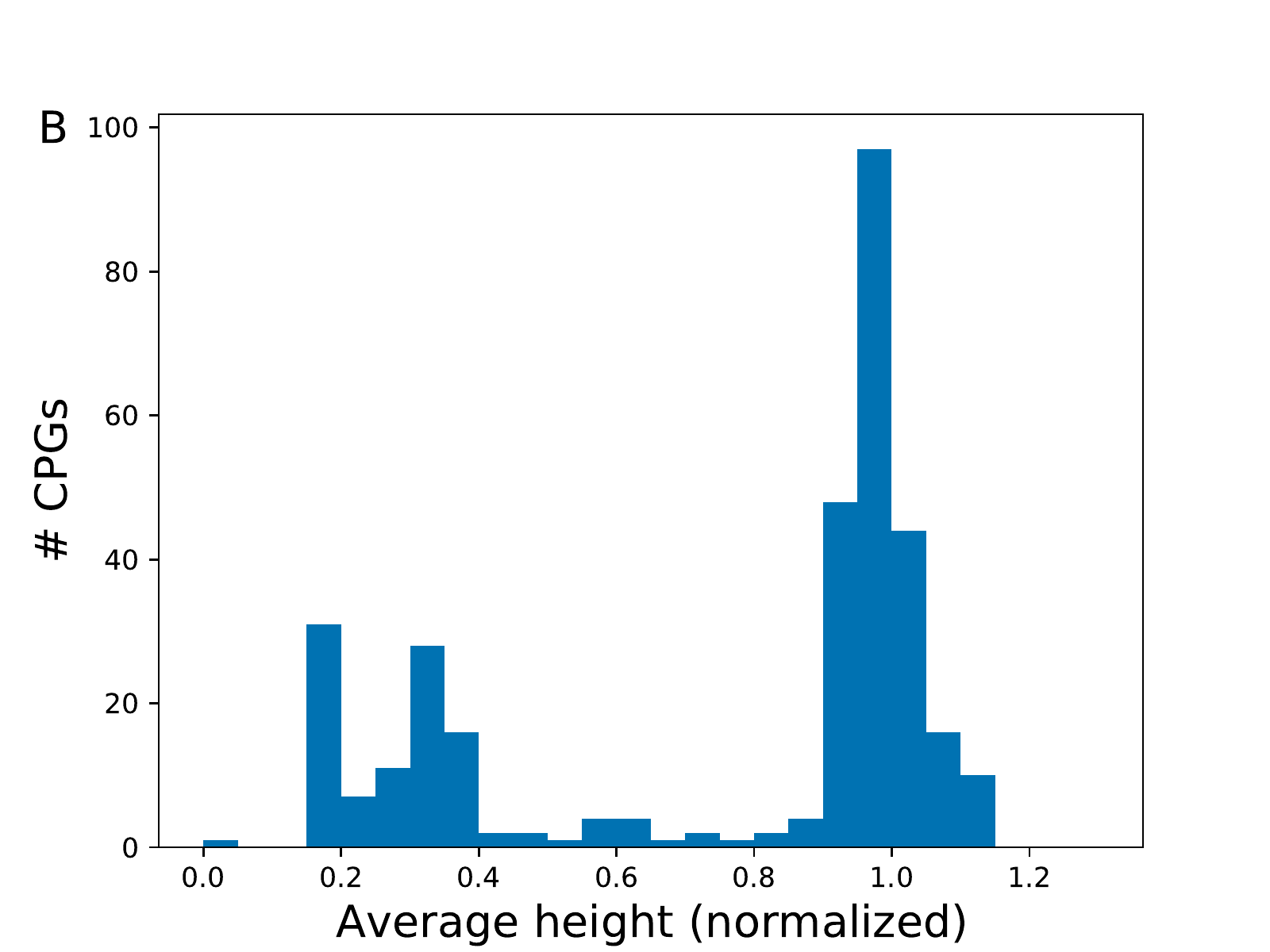}
\caption{Histograms of average height during one evaluation for (A) all normal quadruped CPGs and (B) all short quadruped CPGs, evaluated at $I_{DC}=0.5, \theta_C=0.016$.}\label{fig:height}
\end{center}
\end{figure}

The location of the highest cluster of average height was determined to be above $0.75$, and these CPGs were judged to be upright for the entire trial (see Figure \ref{fig:height}). The cluster of lower average heights corresponds to crawling with knee joints touching the ground, while in between were upright for only part of the trial. All CPGs below $0.75$ average height were excluded from further analysis.

\begin{figure}[t]
\begin{center}
\includegraphics[width=8.5cm]{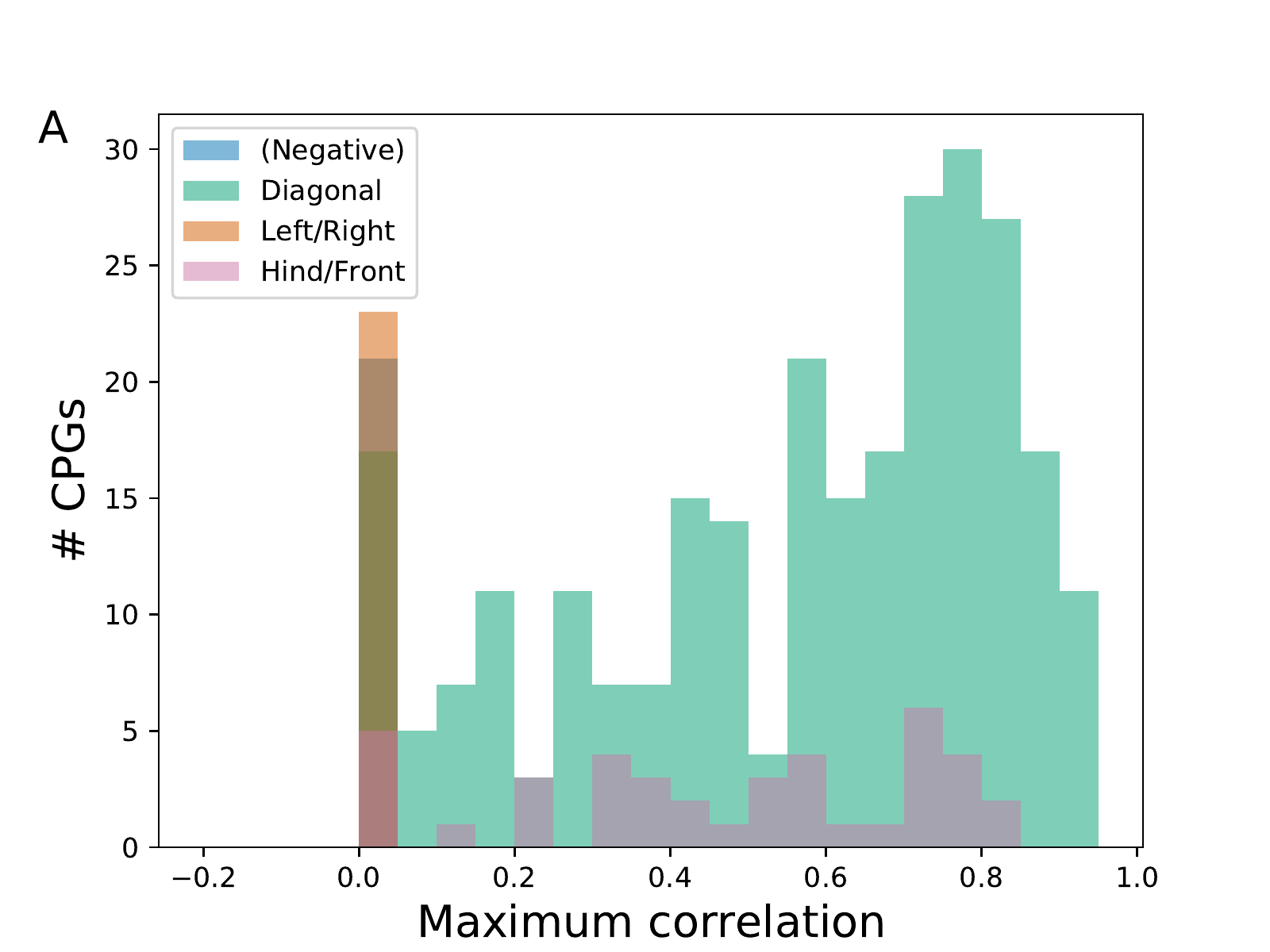}
\includegraphics[width=8.5cm]{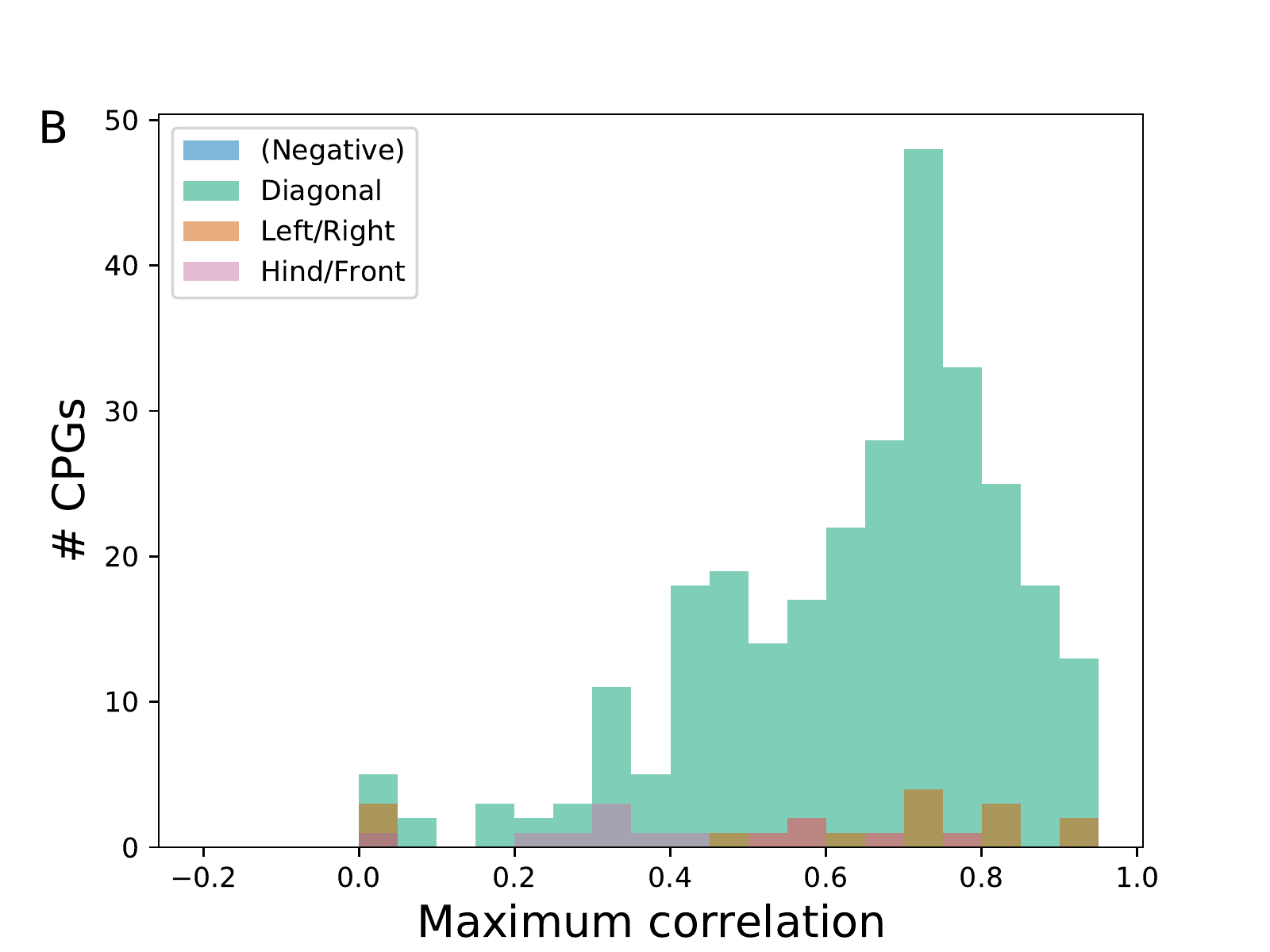}
\caption{Histograms of maximum correlation, sorted by the most correlated pair of limbs, for (A) all normal quadruped CPGs and (B) all short quadruped CPGs, evaluated at $I_{DC}=0.5, \theta_C=0.016$.}\label{fig:corr}
\end{center}
\end{figure}

The interlimb correlation was determined by the Pearson product-moment correlation over the second half of the evaluation period. Only the leg neuron output was used. A four by four matrix $C_{ij}$ of limbs was set up for the correlation between the $i$th and $j$th limb, and only off-diagonal elements were calculated, with diagonal elements $C_{ii}$ being manually set to zero. Therefore if all pairs had negative correlations, the CPG gait was classified as a walk (see Figure \ref{fig:corr}).

\begin{table}[ht]
\centering
\parbox{.45\linewidth}{
\caption{Parameter ranges for the CPG network.}\label{paramtable}\begin{tabular}{c|c}
Parameter & Value / Range \\
\hline
$t_0$       &  0.052 s\\
$\gamma$    & [0.01,0.1] \\
$a$         & [0.2,2]    \\
$b$         & [0.02,0.2] \\
$\kappa$    & [0.5,5] \\
$u_0$       & [0.1,1] \\
$d_i$       & [-0.9,0.9] \\
$c_i$       & [1.1,2] \\
$w_{ij}$    & [-1.8,1.8] / 0\\
\end{tabular}\vspace{1.8cm}
}
\hspace{1cm}\parbox{.45\linewidth}{
\caption{Parameter ranges for the filter network}\label{filtparamtable}\begin{tabular}{c|c}
Parameter & Value / Range \\
\hline
$t_0$       &   0.052 s\\
$\gamma$    &   0.03 \\
$a$         &   2    \\
$b$         &   0.3 \\
$\kappa$    &   4 \\
$u_0$       &   1 \\
$d_i$       &   0 \\
$\tau_0$    &   0.15 \\
$\Gamma$    &   [0.05,0.55] \\
$c_i$       &   [2,2.5] \\
$G_{i}$    &   [-1,1] \\
$w_{ij}$    &   [-1.2,0] \\
$M_{ij}$    &   [-10,10] \\
\end{tabular}
}
\end{table}
\begin{table}[hb]
\centering
\parbox{.4\linewidth}{
\caption{Parameter ranges for the joint activation and sensory feedback. All angles ($\theta$) are in degrees.}\label{quadtable}\begin{tabular}{l|l}
Parameter & Value / Range \\
\hline
$\theta_{0,\mathrm{hip}}$   & [2.7, 27] \\
$\theta_{0,\mathrm{leg}}$   & [4.5, 45] \\
$\theta_{0,\mathrm{knee}}$   & -[7.2, -72] \\
$\theta_{\mathrm{lim,leg}}$   & 90 \\
$\theta_{\mathrm{lim,knee}}$   & 90 \\
$A$         & [0.005, 0.05]   \\
$B$         & [0.005, 0.05]   \\
$q_\mathrm{A,front}$         & [-0.45, 0.45]   \\
$q_\mathrm{B,front}$         & [-0.45, 0.45]   \\
$q_\mathrm{A,side}$         & [-0.45, 0.45]   \\
$q_\mathrm{B,side}$         & [-0.45, 0.45]   
\end{tabular}}\hspace{1cm}
\parbox{.4\linewidth}{
\caption{NSGA-III parameters for each stage of evolution.}\label{evoparams}\begin{tabular}{l|c|c}
 & CPG & Filter \\
\hline
N alleles & 32 & 62 \\
N individuals & 168 & 92 \\
N partitions & 8 & 12 \\
N objectives & 4 & 3 \\
N generations & 200 & 150 \\
$p_c$ & $0.7$ & $1.0$ \\
$p_m$ & $0.05$ & $0.05$ 
\end{tabular}
\vspace{2cm}
}
\end{table}

\subsection*{Statistics}

\begin{figure}[t]
\begin{center}
\includegraphics[width=8.5cm]{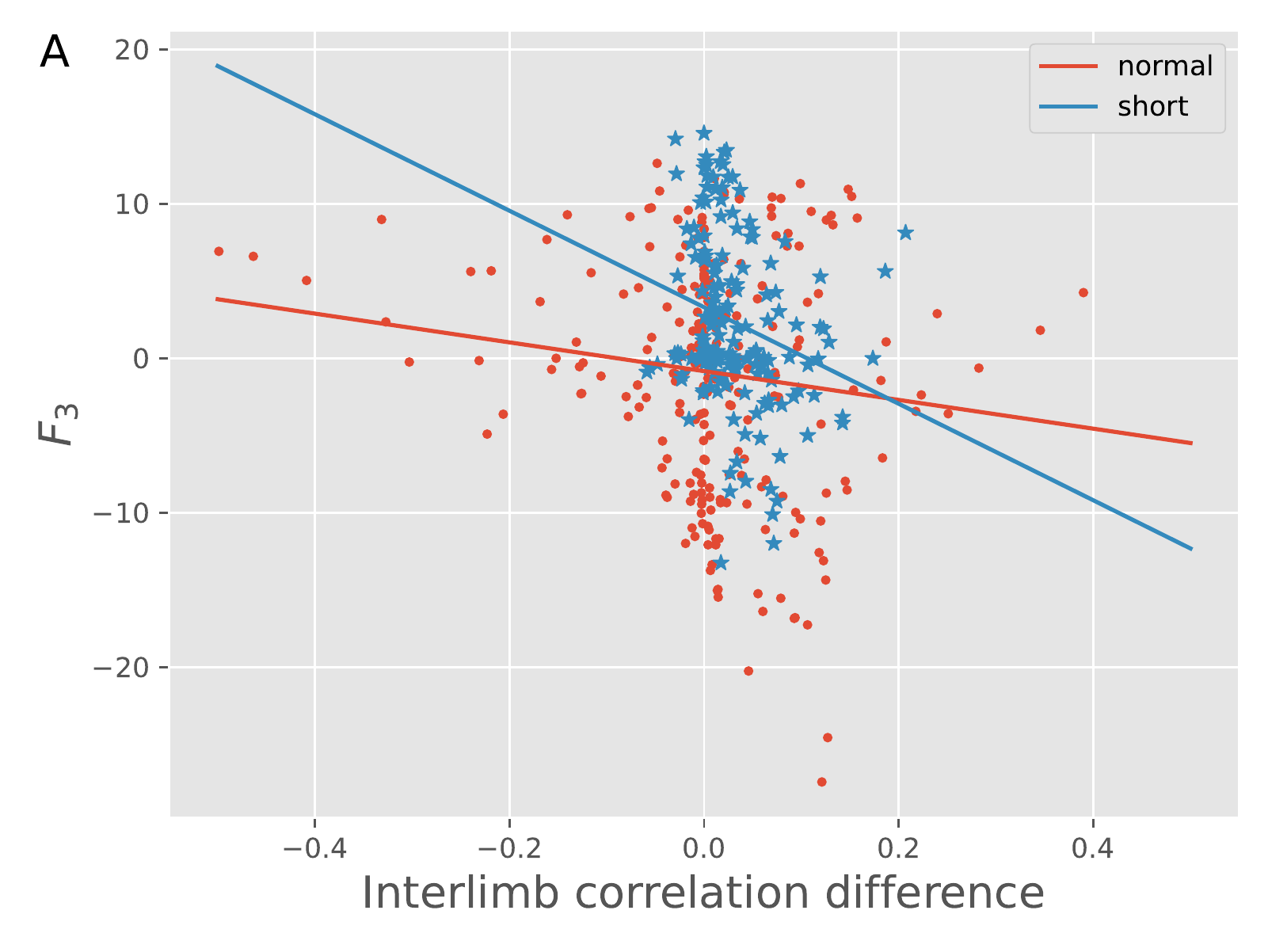}
\includegraphics[width=8.5cm]{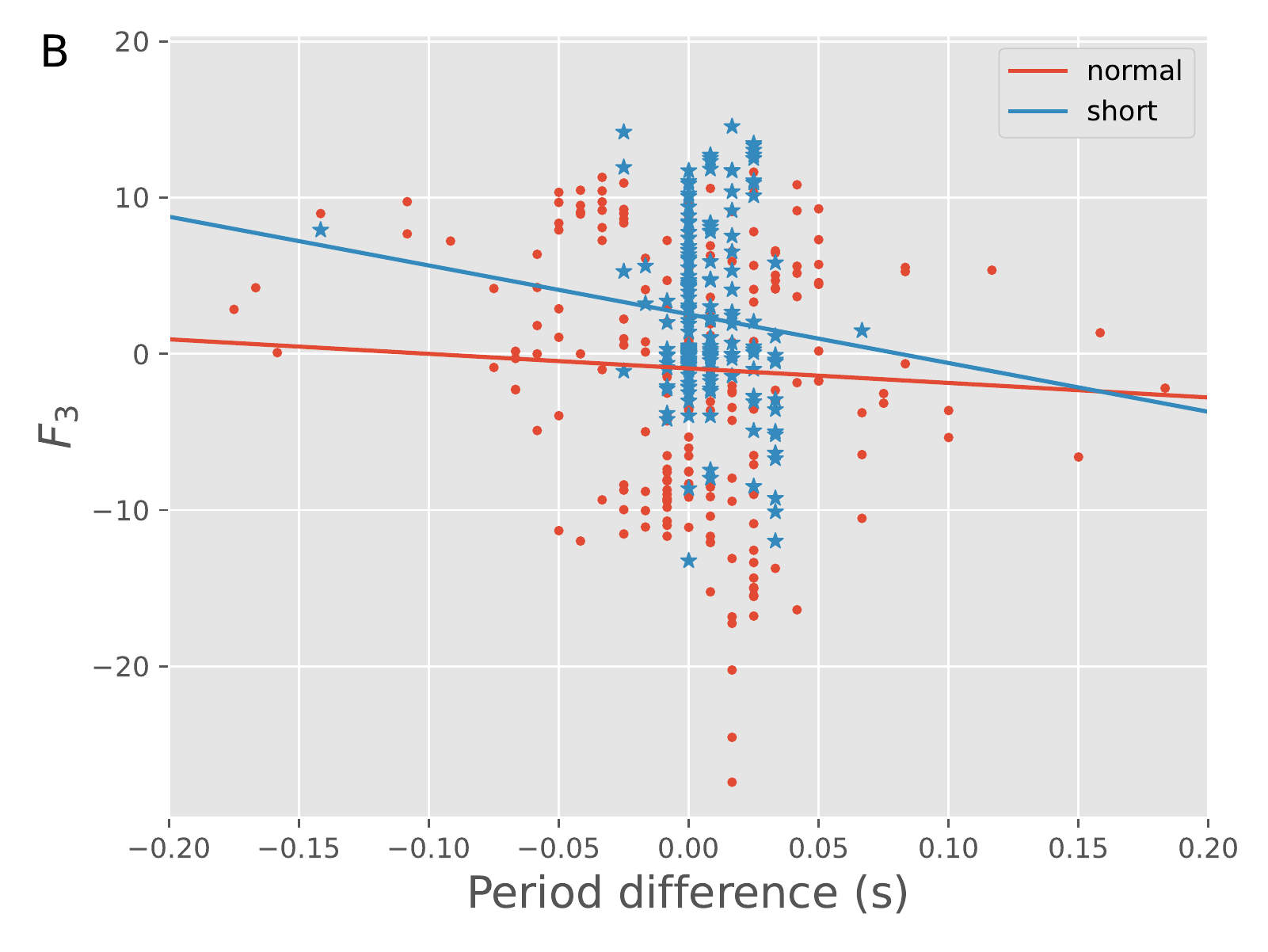}
\end{center}
\caption{$F_3$ as a function CPG variability, for all upright robots. Shown are scatter plots of $F_3$ against \textbf{(A)} the change in the maximum inter-limb correlation, and \textbf{(B)} the change in period, when increasing $I_{DC}$ from $0.5$ to $1$. Straight lines indicate the intercept and coefficient of best fit from linear mixed-effect models. In both panels the robots are in forward mode ($\theta_C=0.016$).}\label{fig:F3S}
\end{figure}

Statistical tests were performed using the Python {\em statsmodels} package. Figure \ref{fig:F3S} shows the lines of best fit from the linear mixed-effect model for $F_3$. Normality of residuals was tested using the Kolmogorov-Smirnov test.

%For more information on Supplementary Material and for details on the different file types accepted, please see \href{http://home.frontiersin.org/about/author-guidelines#SupplementaryMaterial}{the Supplementary Material section} of the Author Guidelines.

%Figures, tables, and images will be published under a Creative Commons CC-BY licence and permission must be obtained for use of copyrighted material from other sources (including re-published/adapted/modified/partial figures and images from the internet). It is the responsibility of the authors to acquire the licenses, to follow any citation instructions requested by third-party rights holders, and cover any supplementary charges.

%% Figures, tables, and images will be published under a Creative Commons CC-BY licence and permission must be obtained for use of copyrighted material from other sources (including re-published/adapted/modified/partial figures and images from the internet). It is the responsibility of the authors to acquire the licenses, to follow any citation instructions requested by third-party rights holders, and cover any supplementary charges.

%\subsection{Figures}

%%% There is no need for adding the file termination, as long as you indicate where the file is saved. In the examples below the files (logo1.eps and logos.eps) are in the Frontiers LaTeX folder
%%% If using *.tif files convert them to .jpg or .png
%%%  NB logo1.eps is required in the path in order to correctly compile front page header %%%

\end{document}